\documentclass[preprint,12pt,authoryear]{elsarticle}

\usepackage{lineno,amsmath, amsthm, amscd,bm, amsfonts,longtable, amssymb, graphicx,multirow,geometry,rotating}
\usepackage[colorlinks]{hyperref}
\usepackage{blindtext}

\usepackage{booktabs}
\usepackage{subcaption,graphicx,placeins}
\usepackage[usenames, dvipsnames]{color}
\usepackage[british]{babel}
\usepackage{lipsum,array, afterpage,pdflscape,booktabs}
\modulolinenumbers[5]
\makeatletter \oddsidemargin.9375in \evensidemargin \oddsidemargin
\theoremstyle{definition}

\theoremstyle{remark}

\numberwithin{equation}{section}

\journal{A}
\allowdisplaybreaks
\begin{document}

\begin{frontmatter}
\title{Addressing Challenges in Time Series Forecasting: A Comprehensive Comparison of Machine Learning Techniques}
\author{Seyedeh Azadeh Fallah Mortezanejad}
\author{Ruochen Wang}
\address{
School of Automotive and Traffic Engineering, Jiangsu University, Zhenjiang, Jiangsu, China.\\
$^*$ Contact Ruochen Wang Email: wrc@ujs.edu.cn }

\begin{abstract}
The explosion of Time Series (TS) data, driven by advancements in technology, necessitates sophisticated analytical methods. Modern management systems increasingly rely on analyzing this data, highlighting the importance of efficient processing techniques. State-of-the-art Machine Learning (ML) approaches for TS analysis and forecasting are becoming prevalent. This paper briefly describes and compiles suitable algorithms for TS regression task. We compare these algorithms against each other and the classic ARIMA method using diverse datasets: complete data, data with outliers, and data with missing values. The focus is on forecasting accuracy, particularly for long-term predictions. This research aids in selecting the most appropriate algorithm based on forecasting needs and data characteristics.
\end{abstract}


\end{frontmatter}

\section{Introduction}
TS is a collection of data points indexed in chronological order. The time
intervals between these data points are typically consistent, such as hourly,
daily, monthly, or yearly. Multivariate TS involves multiple TS variables
measured over the same time period, where the variables are likely
interrelated.

The two most traditional TS datasets are weather data and stock prices.
Hourly or daily measurements of temperature, rainfall, wind speed, and air humidity recorded at a weather station assist district management officials in making critical decisions regarding development, flood prevention, and understanding climate changes. For instance, the quality of seasonal agricultural products is directly affected by climatic conditions.
Daily opening and closing prices of a particular stock are useful for shareholders to predict future prices, identify trading opportunities, and manage risks.
Other examples of TS data include website traffic, sales data, energy consumption, sensor data, medical data, social media activity metrics, economic indicators, and more.
Each of these data points plays an important role in the management of various trades and industries, enhancing the quality of their services and products.

Given the significant advancements in various sectors, multivariate TS models have crucial roles in enhancing performance. Examples of multivariate TS datasets include simultaneously tracking multiple stock prices, measuring temperature, humidity, wind speed, and precipitation at a weather station, monitoring traffic volume on various roads in a city, tracking power generation from multiple sources alongside electricity demand across different regions, and monitoring multiple vital signs for a patient, among others. \cite{luo2024moderntcn} addressed the low performance of convolutional models in TS analysis. They proposed Modern Temporal Convolutional Network (ModernTCN) with modifications tailored for TS tasks. ModernTCN achieves state-of-the-art performance on five benchmark tasks while maintaining the efficiency of convolutional models, exhibiting a superior balance of performance and efficiency compared to transformer and multi-layer perceptron based approaches.

\cite{torres2021deep} reviewed Deep Learning (DL) architectures, including
Feed-Forward Network (FFN)s, Convolutional Neural Network (CNN)s, Recurrent
Neural Network (RNN)s such as Elman, Long Short-Term Memory (LSTM), Gated
Recurrent Units (GRUs), and bidirection for TS forecasting. They discussed
the advantages and limitations of these architectures, hyper-parameter
tuning, suitable frameworks, and identified research gaps across various
application domains. \cite{lim2021time} surveyed encoder-decoder DL
architectures for one-step-ahead and multi-horizon TS weather forecasting. It
highlighted how these models incorporate temporal information and examined
recent hybrid models that combine statistical and Neural Network (NN) approaches.

Since the importance of TS data has grown, many libraries in MATLAB, Python, and other programming languages have been developed to implement state-of-the-art methods for analyzing TS data.
\cite{tavenard2020tslearn} introduced the tslearn library, a Python package designed for ML tasks such as clustering, classification, and regression specifically tailored for TS data.
PyTS is an open-source Python package for TS classification, as established in \cite{faouzi2020pyts}. It provides a variety of algorithms, preprocessing tools, and utilities for loading datasets, all built on standard scientific Python libraries, and is available on GitHub.
\cite{hoffmann2021deeptime} presented another Python library designed for estimating dynamic models from TS data. This library offers a range of methods, including traditional linear approaches as well as DL techniques, and it provides comprehensive analytical capabilities for calculating thermodynamic, kinetic, and dynamical quantities.
\cite{flood2021entropyhub} introduced EntropyHub, an open-source toolkit available in MATLAB, Python, and Julia, which offers over forty functions for calculating various entropy measures, including cross-entropy, multi-scale entropy, and bidimensional entropy for TS analysis. The toolkit aims to consolidate existing entropy methods into a single, easily accessible, and reproducible resource.
\cite{schirmer2024pydts} explored TS modeling, emphasizing five key applications: denoising, forecasting, nonlinear transient modeling, outlier detection, and degradation modeling. The study categorizes modeling approaches into statistical, linear algebra, and ML methods while reviewing relevant literature. A key contribution of the article is the introduction of PyDTS, a Python-based toolkit that integrates various TS modeling techniques, complete with practical examples and benchmark comparisons across diverse datasets.

In this paper, we compile various ML techniques that are suitable for TS regression tasks. In this regard, we also include the traditional ARIMA model to demonstrate its limitations in forecasting the future of TS datasets. This paper highlights the effectiveness of ML in solving everyday problems with satisfactory accuracy. By considering ill-posed TS datasets as examples, we observe the high performance of these algorithms in both long-term and short-term predictions, without the extensive data preparation required by ARIMA and many other traditional models.

The contribution of this paper are outlined as follows: In Section \ref{S2},
we provide a brief explanation of the TS concept and forecasting using ARIMA
models. In Section \ref{S3}, we discuss feature selection relevant to TS
analysis. In Section \ref{S4}, we compile algorithms suitable for TS
regression tasks. In Section \ref{S5}, we examine algorithms appropriate for
handling outliers data. In Section \ref{S6}, we address the issue of missing
data and the corresponding algorithms. In Section \ref{S7}, we present the
results of three case studies. Finally, in Section \ref{S8}, we share the
insights derived from this paper.
\section{Traditional TS Concepts}\label{S2}
A fundamental concept in TS analysis is stationarity. A stationary TS exhibits consistent statistical properties, such as mean, variance, and autocorrelation, over time.
The simplest method to check for stationarity is the TS plot. In this plot, several key features must be examined, including trends, seasonal components, and any malformed behaviors such as sharp points and outliers.
For more information, let's get into some mathematical definitions.
A TS model represents a specific joint distribution, or at the very least, the means and covariances of a randomly ordered chronological variable $\{X_t\}$, with observations denoted as $\{x_t\}$.
The mean, variance, and covariance functions for a TS $ \{X_t\} $ with $E(X_t)^2<\infty$, are defined as follows:
\begin{align*}
  \mu_X(t) & =E(X_t), \\
  Var_X(t) & =E[ (X_t-\mu_X(t))^2], \\
  Cov(X_t,X_s) & = E[ (X_t-\mu_X(t))(X_s-\mu_X(s)) ].
\end{align*}
A TS $\{X_t\}$ is said to be strictly stationary if the joint distribution of $(X_{t_1},\cdots,X_{t_k})$ is the same as the joint distribution of $(X_{t_1+h},\cdots,X_{t_k+h})$ for all integers $h$ and $k\geq 1$.
Also, $\{X_t\}$ is considered as a weakly stationary TS if it satisfies these conditions:
\begin{itemize}
  \item $\mu_X(t)$ and $Var_X(t)$ are independent of time $t$.
  \item $Cov(X_{t+h},X_t)$ is independent of $t$ for every integer $h$.
\end{itemize}
In practice, the definition of weak stationarity is commonly used and referred to simply as stationarity.
A TS is considered non-stationary if any of these conditions are violated. Common indicators of non-stationarity include:
\begin{itemize}
  \item Trending: the mean of the series changes systematically over time;
  \item Seasonality: the series exhibits periodic fluctuations;
  \item Heteroskedasticity: the variance of the series changes over time.
\end{itemize}
In traditional modeling and forecasting methods, the first and most crucial step is to transform a non-stationary TS into a stationary one.
The presence of a trend is indicated by a consistent increase or decrease in values over time.
A key seasonality characteristic is the recurrence of specific patterns at regular intervals.
Heteroskedasticity can be identified by visualizing variations in the frequency of patterns across different periods.
Common techniques to address these issues include first-order differencing, defined as $ Y_t = X_t - X_{t-1} $, which removes trends; seasonal differencing of lag $h$, represented as $ Z_t = Y_t - Y_{t-h} $, where $ h$ corresponds to the fluctuation period; and log transformation to address heteroskedasticity.
There are some other methods like Seasonal-Trend decomposition using Locally Estimated Scatterplot Smoothing (LOESS) 
that help finding the trends and seasonal components. STL uses locally fitted regression models to decompose a TS into trend, seasonal, and the remaining components.
Furthermore, there are several statistical tests, such as the Augmented Dickey-Fuller
, Kwiatkowski-Phillips-Schmidt-Shin
, and Mann-Kendall tests, used to analyze unit roots and trends.
These can be helpful for beginners in diagnosing non-stationarity in TS analysis.

There are several TS models to forecast a stationary TS such as Moving Average ($MA$), Autoregressive ($AR$), Autoregressive Integrated Moving Average ($ARIMA$), Seasonal Autoregressive Integrated Moving Average ($SARIMA$).
An $MA(q)$ model represents a TS with a weighted average of current and past error terms like $X_t=\mu+\epsilon_t+\theta_1\epsilon_{t-1}+\cdots + \theta_q\epsilon_{t-q}$, where random error terms $\epsilon_t$s are independent and identically distributed ($i.i.d.$) and $\theta_i$s are the model parameters.
An $AR(p)$ models a TS as a linear combination of its past values and an error term $X_t=\phi_0+\phi_1X_{t-1}+\cdots + \phi_pX_{t-p}+\epsilon_t$, where $\phi_i$s are the model parameters.
An $ARIMA(p, d, q)$ model combines $AR$ and $MA$ components and incorporates differencing $d$. The value of $d$ in ARIMA represents the number of differences needed to achieve stationarity.
A $SARIMA(p, d, q)(P, D, Q)$ model extends $ARIMA$ to account for seasonality with period $s$. The capital letters $(P, D, Q)$ represent the seasonal $AR$ and $MA$ components.

Choosing the appropriate model necessitates careful consideration of the data's characteristics, including trends, seasonality, and the Autocorrelation Function (ACF) as well as the Partial Autocorrelation Function (PACF).
Generally, the values of $q$ and $p$ can be approximately determined based on the number of significant ACF and PACF lags, respectively.
Model selection generally involves conducting diagnostic checks to ensure a proper fit and to evaluate the model's residuals for stationarity. If stationarity is not attained through preprocessing, directly applying these models may result in inaccurate or unreliable forecasts, \cite{brockwell2002introduction}.
The traditional models perform effectively with data that exhibits significant values in the ACF and PACF only for the most recent lags. Consequently, these models may struggle to manage data with large values at higher lags.
Additionally, TS data with persistent patterns are intractable for making predictions using the classical models.

\section{TS Feature Engineering for ML Algorithms}\label{S3}
In this paper, we aim to implement several ML algorithms for TS analysis. Traditional methods often struggle with complex data patterns and require preprocessing to achieve stationarity in TS data. In contrast, ML methods are known for their robustness across various tasks, providing several advantages over traditional TS forecasting methods.
ML algorithms can identify complex patterns and nonlinear relationships within data that traditional linear models may overlook. TS data frequently displays nonlinear patterns, which makes ML methods particularly well-suited for the analysis.
The incorporation of multiple external predictor variables is straightforward in ML, enhancing forecasting capabilities, which is not as simple with traditional methods.
ML algorithms are flexible and can adapt to changing patterns in data. They can be periodically retrained with new data to maintain their accuracy.

However, applying ML algorithms to TS data requires careful feature engineering. Raw TS data is often unsuitable as direct input to most ML algorithms. Feature engineering transforms the raw data into a format more amenable to ML model. This process involves creating new features from the existing data that might improve model performance.

A common approach for preparing TS data for ML models is to create lagged features. This involves creating input vectors comprising previous time steps' values. For example, if we want to predict the value at time $t$, and use a lag of $h$, the input and output are $(X_{t-h},\cdots,X_{t-1})$ and $X_t$, respectively.
Another method involves calculating rolling statistics, including rolling means, standard deviations, minimums, maximums, and medians over a specified window size (e.g., the rolling mean over the past $ h $ days). These features effectively capture trends and variations within the data. Additionally, there are several other features that can be selected for a specific problem, such as time-based features, seasonal features, trend indicator, and external predictor variables.
Feature selection techniques assist in identifying the features that have a significant impact on the output.
Additionally, data normalization can enable the machine to learn more quickly than if the input is provided in its original scale. Splitting the data into three sets—training, validation, and testing—facilitates a more effective evaluation of the machine's performance, \cite{joseph2022modern}.
\section{ML Algorithms for Sequential Data}\label{S4}
In this section, we aim to comprehensively collect the ML algorithms that are suitable for TS forecasting and provide a brief explanation of each.
A concise overview is provided by highlighting its strengths, weaknesses, and suitability for different types of TS data. The algorithms are categorized for ease of understanding.
\subsection{Recurrent Neural Networks (RNNs)}
RNNs are a subset of NNs specifically designed for processing sequential or temporal data, enabling them to effectively capture nonlinear short-term dependencies effectively.
RNNs are widely used in processing text, speech, and TS data because they can handle information across multiple time steps. RNN neurons transmit feedback signals to one another, allowing them to maintain a kind of memory for past inputs. 
RNN architectures are distinguished by its cyclical connections, which allow information to persist across time steps.
RNNs incorporate loops that feed the output of a layer back into itself, which creates a form of memory.
This enables the network to take past inputs into account when processing current inputs.
However, standard RNN architectures suffer from the vanishing gradient problem, limiting their ability to learn long-term dependencies, \cite{tsantekidis2022recurrent}. To mitigate this, several variations have been developed.
    \subsubsection{Elman RNN}
    Elman (Vanilla) RNN is the simplest type of RNN with a context layer that stores a copy of the previous hidden state. While effective for shorter sequences, they still struggle with very long-term dependencies.
    \subsubsection{Long Short-Term Memory (LSTM) Network}
    LSTM networks are more complex than standard RNNs but are significantly better at capturing long-range dependencies.
    LSTMs address the vanishing gradient problem through a sophisticated cell state mechanism, \cite{noh2021analysis}. This allows them to learn long-range dependencies more effectively than standard RNN or Elman networks. The gating mechanisms, including input, forget, and output gates, regulate the flow of information into and out of the cell state. They use sigmoid activations to produce values between $0$ and $1$, controlling the flow of information.
    \cite{lindemann2021survey} published a survey about different types of RNN, particularly focusing on LSTM for TS.
    LSTM networks are designed for sequential data, such as TS, but it is not efficient in extracting spatial relationships. In this context, a CNN is preferable. We discuss about CNN later in this section.
    \subsubsection{Gated Recurrent Unit (GRU) Network}
    GRUs
    are a simplified version of LSTM networks. The forget and input gates are combined into a single update gate and merged the cell state with the hidden state. This design reduces the number of parameters, making GRUs computationally less expensive and often easier to train than LSTMs, while still delivering strong performance across various applications.
    \cite{mateus2021comparing} compared the performance of LSTM networks and GRU in a case study where traditional models such as ARIMA and SARIMA struggled to capture the stochastic nature of the data. As anticipated, the performance of the GRU outperformed that of the LSTM in their analysis.
    \subsubsection{Bidirectional RNN}
     Bidirectional RNNs process input sequences in positive and negative time directions known as forward-backward states. By combining the information from both directions, bidirectional RNNs can capture contextual information from both the past and the future, leading to improved performance in tasks where understanding the entire sequence is crucial. This is particularly beneficial for TS forecasting, where knowledge of future data points within a limited window can enhance predictions.
     The networks can be implemented using two RNN layers, including LSTMs and GRUs.
     The concept behind bidirectional RNNs is to separate the state neurons into forward-backward states, with the inputs of each component remaining unconnected to the other.

     These models can process data in both forward-backward sequential directions, increasing their ability to understand and interpret textual information more effectively than traditional RNNs, \cite{hindarto2023comparison}.
     The model's architecture exhibits symmetry in its directional flow, which is identified as a drawback in textual problems. This limitation arises because word positions are determined by grammar rather than by temporal context. Another issue with bidirectional RNNs is the concatenation of the forward-backward states without any specific considerations, \cite{rozenberg2021asymmetrical}.
     \cite{zhao2020double} introduced an enhanced hybrid algorithm that combines bidirectional LSTM networks and CNNs to effectively process sequential and spatial data.
    \subsubsection{Deep RNN}
    Instead of using a single or two recurrent layers, deep RNNs stack multiple recurrent layers on top of each other. This allows the network to learn hierarchical representations of the input sequence, capturing both low-level and high-level features. Deep RNNs can be more powerful than shallow RNNs, particularly for complex TS with intricate patterns. However, they are also more computationally expensive and require more data for effective training.
    There are several papers worked on deep RNNs for different tasks. For instance, \cite{ahn2021deep} developed deep RNNs for short-term forecasting of fluctuating photovoltaic power output. The model achieved high accuracy for forecasts made $5$ and $15$ minutes in advance. However, the accuracy decreased slightly for longer forecasts ranging from $1$ to $3$ hours. The deep RNN-based forecasting algorithm demonstrated superior accuracy in predicting short-term photovoltaic power generation.
    \subsubsection{Deep Autoregressive RNN (DeepAR)}
    DeepAR, \cite{salinas2020deepar}, is a probabilistic forecasting model developed by Amazon. It is designed for TS forecasting, particularly when dealing with many independent TS. DeepAR leverages LSTMs, to capture temporal dependencies within each TS. Crucially, it is a probabilistic model, meaning it produces not just a point forecast but also a probability distribution over future values. This allows for quantifying the uncertainty in the forecast, \cite{salinas2020deepar}.

    DeepAR uses an autoregressive RNN architecture. This means it predicts future values based on past values of the same TS. The LSTM processes the historical data, and the model learns to map past observations to a probability distribution over future values. A significant feature is its ability to handle many TS simultaneously by learning shared patterns across them. This is achieved by using an embedding layer that represents each TS as a vector.

    Like many other algorithms, it involves costly computations and requires tuning of hyper-parameters for improved performance. It struggles with very long-range and noisy data, similar to LSTM, and presents challenges in interpretation.

\subsection{Tree-Based Ensemble Methods}
Tree-based ensemble methods have become increasingly popular for TS forecasting due to their ability to handle nonlinear relationships, automatically capture feature interactions, and provide robust performance across diverse datasets, \cite{rady2021time} and \cite{mir2022anomalies}. These methods build multiple decision trees and combine their predictions to improve accuracy and reduce over-fitting.

The strength of these models is in their ability to capture complex patterns commonly found in temporal data, such as seasonality, trends, and cyclical behaviour. Common algorithms used in this context include Random Forest (RF), Extreme Gradient Boosting (XGBoost), and Light Gradient-Boosting (LightGBM). For regression tasks in TS forecasting, the output of the ensemble is a continuous value that represents the predicted future value of the TS.
Feature engineering such as lagged values, rolling statistics, and external regressors, plays a crucial role in enhancing forecast accuracy. 
    \subsubsection{Random Forest (RF)}
RF is an ensemble learning method used for both classification and regression tasks.
RF constructs multiple decision trees during training to enhance accuracy and control over-fitting.
Its outputs represent the prediction mode and mean of the individual trees for classification and regression tasks, respectively.
A random subset of features is selected when splitting nodes to further reduce correlation among trees and enhance model robustness.
RF is a rapid algorithm that effectively manages noise and outliers, outperforming many other algorithms. Additionally, it does not require data scaling.
By increasing the number of trees, over-fitting may occur, and inadequate features can result in suboptimal performance.
    \subsubsection{Extreme Gradient Boosting (XGBoost) Machine}
    XGBoost
    is another highly popular gradient boosting library known for its performance and robustness.
 It builds an ensemble of decision trees sequentially, with each new tree learning from the mistakes of the previous ones. Each tree is built by splitting nodes to maximize a gain function. This function considers both the improvement in accuracy and $L1$ and $L2$ regularization to avoid over-fitting. XGBoost cleverly handles missing data and uses parallel processing for faster tree construction. However, it is computationally expensive, memory-intensive, requires careful tuning of its hyper-parameters.
 There are several applications of XGBoost.
    \subsubsection{Light Gradient-Boosting (LightGBM) Machine}
    LightGBM
    is a gradient boosting framework that stands out due to its speed and efficiency, particularly for dealing with large TS datasets. It achieves this through several key optimizations in its algorithm structure.
    LightGBM is the optimized version of XGBoost algorithm.
    The training data is partitioned into several bins to approximate the data distribution by histograms. This significantly reduces the memory usage and speeds up the training process.
    Each leaf node stores a histogram containing the sum of gradients and the sum of Hessians for each data point.

    Trees are constructed iteratively. At each iteration, LightGBM evaluates the gain for each possible split for each leaf and selects the leaf with the largest gain at each step, leading to deeper trees with potentially higher accuracy. However, this can also lead to over-fitting if not carefully controlled.
    To speed up the training process and reduce the impact of data instances with smaller gradients, Gradient-based One-Side Sampling (GOSS) technique randomly samples a subset of data with larger gradients and keeps all data with smaller gradients. This focuses computational effort on the more informative data points.
    To reduce the dimensionality of the data, exclusive features with minimal overlap are gathered.
    LightGBM incorporates $L1$ and $L2$ regularization to prevent over-fitting.
    LightGBM requires careful hyper-parameter tuning to achieve optimal performance.

    So many papers used this algorithm for different TS problems. \cite{cao2023greenhouse} proposed a LightGBM tree to predict greenhouse temperature TS. LightGBM significantly outperformed other models, such as support vector machine, radial basis function NN, back-propagation NN, and multiple linear regression model, in terms of MSE and R-squared.
\subsection{Specialized TS Models}
This subsection explores specialized TS models that offer distinct advantages over general-purpose algorithms. These advantages include effectively managing strong seasonality and trend components, handling missing data, incorporating multiple exogenous variables, and accounting for temporal dependencies.
    \subsubsection{Facebook Prophet}
    Facebook Prophet
    is a TS forecasting tool designed for business applications, particularly those with strong seasonality and trend components, as it was used in \cite{daraghmeh2021time}. It is known for its ease of use and ability to handle missing data and outliers.
    It decomposes the TS into three components: trends, seasonalities, and vacation irregularities.
    The trend component represents long-term developments, while the seasonal component accounts for periodic variations.

    A nonlinear function captures the overall direction of the TS trends.
    The Prophet utilizes the Fourier series for specified periods, such as daily, weekly, or yearly, to create an effective model that incorporates periodic effects.
    Prophet is an effective tool for TS analysis, accommodating multiple levels of seasonality, which makes it suitable for a variety of data types.
    The model can automatically identify and learn these patterns.
    The effects of holidays and other special events are presented as a list of dates.
    Additionally, it is equipped with external regressors that can impact the TS.
Tuning certain hyper-parameters can help achieve optimal solutions. Furthermore, Prophet is intended for long-term TS analysis, rendering it ineffective for short-term predictions.

    \subsubsection{Wavelet-Based Transformation (WBT)}
Wavelets are mathematical functions that can transform data into different scales or resolutions, and they are commonly used for TS decomposition. Unlike traditional Fourier transforms, which decompose signals into sine and cosine functions, wavelets can capture both frequency and location information, making them particularly useful for non-stationary TS.
TS data can be decomposed into various components, such as trends, seasonality, and noise, using WBTs. The coefficients obtained from WBTs can serve as features for ML models.

However, wavelet analysis can be mathematically complex, which may make it challenging to extract useful features efficiently. The performance of WBTs heavily depends on the choice of wavelet function (e.g., Haar, Daubechies, Coiflets). Additionally, wavelet features can lead to over-fitting, particularly if model complexity is not managed appropriately.
Many other factors must be considered when using WBTs, such as parameter tuning and potential information loss during decomposition. While WBTs are effective for non-stationary data, they may struggle with very complex non-stationarity patterns and are susceptible to boundary effects, which can impact forecasting accuracy.
    \subsubsection{Temporal Fusion Transformer (TFT)}
    TFT, \cite{lim2021temporal}, is a DL model designed for multi-horizon forecasting TS data by selecting features that exhibit strong interactions with the target variable.
    Its architecture incorporates LSTM networks to effectively capture long-term dependencies. Additionally, TFT combines the advantages of attention mechanisms with a thoughtfully constructed architecture that explicitly addresses various aspects of TS data, including known future inputs and static covariates.
    The attention mechanism enhances interpretability by highlighting which past time steps and features are most influential in making predictions. As a result, it provides greater insight compared to many black-box methods.
    Also, gated residual networks in TFT filter out unrelated algorithm nodes to enhance implementation speed.

    TFT can handle various types of features, including numerical and categorical data, to gain deeper insights into the TS.
    Besides all these effective performances, it has high computational costs due to its complex architecture. Additionally, the decision-making process can be quite challenging.
    \cite{lopez2022application} explained TFT and compared its results with those of simple ARIMA and other NNs, such as LSTM, multilayer perceptron, and XGBoost, for day-ahead photovoltaic power forecasting.
    \subsubsection{Neural Basis Expansion Analysis for TS (N-BEATS)}
N-BEATS
is designed for univariate TS and features a dense architecture incorporating forward-backward residual links. It can be applied to a wide range of data structures without requiring any modifications to the architecture.
Each fully connected block learns a distinct component of TS data, such as a specific pattern or trend. These blocks are referred to as basis functions, and their outputs are aggregated to generate the final forecast.
Despite the dense, deep layers, N-BEATS offers fast performance that is user-friendly and achieves high accuracy in predictions.
Although it is designed for TS purposes, it can be effectively applied to non-TS data, \cite{oreshkin2019n}.

The decomposition into basis functions offers a degree of interpretability, enabling an understanding of which components contribute most significantly to the forecast.
The fully connected nature makes it relatively straightforward to scale to longer TS and larger datasets.
Compared to RNNs and LSTMs, this architecture is simpler to implement and train.
It may have difficulty handling very complex or irregular seasonal patterns when compared to specialized models that are specifically designed for these scenarios.
Fully understanding the learned basis functions can be challenging, particularly when dealing with a large number of blocks.
Like many DL models, the performance is sensitive to hyper-parameter tuning.
Finding the optimal configuration may necessitate considerable experimentation.
While some extensions exist, the fundamental N-BEATS architecture does not directly incorporate external regressors or additional features that influence the TS.

    \subsubsection{Neural Hierarchical Interpolation for TS (N-HiTS)}
N-HiTS, \cite{challu2023nhits}, is another DL model designed for TS forecasting, particularly effective for handling long-term dependencies and high-frequency data. Unlike N-BEATS which uses a stack of fully connected blocks, N-HiTS utilizes a hierarchical structure. This structure decomposes the forecasting problem into multiple levels, each responsible for predicting a specific temporal resolution or frequency.
\cite{jeffrey2023development} applied N-HiTS to forecast stock prices using multivariate TS analysis and demonstrated the effectiveness of N-HiTS for short-, medium-, and long-term stock price forecasting.

The decomposition into various levels facilitates the efficient processing of high-frequency TS, which can be computationally intensive for other models.
It naturally offers forecasts at various resolutions, enabling both coarse-grained long-term predictions and fine-grained short-term forecasts.
In scenarios involving complex long-term dependencies and multi-resolution requirements, N-HiTS can outperform simpler methods.
The hierarchical structure adds more complexity than simpler models, such as N-BEATS, resulting in longer training times.
Comprehending the internal mechanisms of each level poses challenges.
As a result of increased complexity, there is a larger hyper-parameter space that requires more careful tuning to achieve optimal performance.
Optimal performance often necessitates a significant amount of data to effectively train the model at each hierarchical level.

\subsection{Other Neural Network Architectures for TS}
In this subsection, we explore various famous algorithms that are utilized for a range of problems, including their application to TS forecasting tasks.
This requires a deep understanding of TS data and the relative features affecting future outcomes. Therefore, the features must be carefully selected for inputs.
    \subsubsection{Deep Feed-Forward Neural Network (DFFNN)}
FFNs are well-known alongside ML tools, which process data in a single pass.
DFFNNs for TS forecasting provide a straightforward approach by utilizing multiple layers of fully connected neurons, making them relatively easy to implement.
Unlike RNNs or specialized architectures such as N-HITS, it does not inherently possess mechanisms for managing long-range dependencies or multi-resolution data. Instead, it relies on the representational power of deep layers to learn complex patterns from the input data. For TS analysis, this often involves preparing the input data as a sequence of lagged values and rolling statistics.

DFFNNs are relatively simple to implement and comprehend compared to more complex architectures.
Although the computational cost increases, they can be scaled to manage large datasets and longer TS.
DFFNNs can be adapted for various forecasting tasks by modifying the input preparation and output layers.
Standard DFFNNs often struggle to capture long-range dependencies. Information from distant past time steps can be lost or diluted as it propagates through the network.
DFFNNs do not inherently provide multi-resolution forecasts; obtaining predictions at different frequencies. This requires the use of separate models or post-processing techniques.
The performance is highly dependent on the preparation of the input data.
With numerous layers and parameters, DFFNNs are susceptible to over-fitting if not properly regularized.
Other drawbacks of traditional training algorithms, such as back-propagation, include slow convergence speed and a tendency to fall into local minima, which can diminish the performance of the classifier or regressor, \cite{ab2021integrating}.
    \subsubsection{Bayesian Neural Network (BNN)}
Bayesian ML algorithms are especially effective in TS forecasting because they can integrate prior beliefs and handle uncertainty in both parameters and predictions. This makes them well-suited for dynamic and noisy environments, such as those found in TS data.
Bayesian methods can effectively model complex relationships and incorporate hierarchical structures.
They tend to be more resilient to over-fitting, particularly when regularization is applied through prior distributions.
Common Bayesian methods for TS forecasting include Bayesian linear regression, Gaussian processes, Bayesian structural TS, Bayesian hierarchical models, dynamic linear models, and BNNs.

Bayesian inference can be computationally intensive and may have slow implementations, especially when dealing with large datasets or complex models.
Setting up Bayesian models can be more complex than traditional methods, as they require a deeper understanding of the underlying probabilistic frameworks.
The selection of prior distributions can significantly impact the results. Inappropriately chosen priors may lead to biased predictions.
Some Bayesian methods struggle to scale effectively with large datasets, which makes them less suitable for very high-dimensional or large-scale TS data.

In this paper, we focus on BNNs.
In BNNs, rather than optimizing a single set of weights, distributions over weights are inferred. This approach enables the model to express uncertainty regarding its predictions. Consequently, BNNs provide a posterior distribution over weights, allowing the model to adapt based on the data it encounters.
    \subsubsection{Convolutional Neural Network (CNN)}
CNNs are primarily recognized for their effectiveness in image processing, particularly when spatial relationships exist between neighbouring points. Consequently, they are applied to TS forecasting. In this context, the TS data is treated as a one-dimensional image, with each data point representing a pixel. The convolutional filters traverse the TS, identifying local patterns and features.

CNNs are exceptional at identifying local patterns and features within TS data, which can be crucial for short-term forecasting.
Compared to fully connected networks of similar capacity, CNNs typically require fewer parameters. This reduction minimizes the risk of over-fitting and accelerates the training process.
Many well-established libraries offer efficient implementations of CNNs, making them user-friendly.
Unlike some recurrent models, CNNs can be adapted to handle TS data of varying lengths by utilizing padding techniques.

    \subsubsection{Temporal Convolutional Network (TCN)}
TCNs are a specialized type of CNN designed explicitly for sequential data, such as TS. They address some of the limitations of standard CNNs in handling long-range dependencies by employing techniques like dilated convolutions and residual connections.
Dilated convolutions allow TCNs to have a large receptive field without an excessive number of parameters or computational costs, enabling them to capture long-range dependencies more effectively than standard CNNs.

Unlike RNNs, TCNs can be parallelized during training and inference, leading to significantly faster processing times, especially for long sequences.
The residual connections help stabilize the training process, mitigating the vanishing gradient problem often encountered in deep networks.
Similar to CNNs, TCNs can handle variable-length sequences with appropriate padding.
The optimal configuration of dilated convolution rates and filter sizes can significantly impact performance and require careful tuning.
Although more efficient than RNNs, processing extremely long sequences can still be computationally demanding, especially with a large number of layers and wide filters.
In certain scenarios, other specialized architectures like transformers or N-HITS might still outperform TCNs, particularly for very complex TS.

\subsection{Advanced Techniques}
In addition to all suitable algorithms for TS forecasting, including those specifically designed for this purpose, we include a subsection on three additional models that are recognized as advanced methods in ML for various types of data. Here, our focus is on the aspects related to TS analysis.
    \subsubsection{Short-Time Fourier Transformation (STFT)}
STFT is a method for analyzing the frequency content of non-stationary signals or TS.
It divides data into shorter overlapping segments and applies the Fourier transform to each one, allowing for the examination of how frequency components change over time.
The Fourier transform converts the time-domain signal into the frequency domain to extract features from the resulting spectrogram.
These invisible features can be utilized in other forecasting models, such as LSTM and GRU.
This makes STFT particularly useful in TS forecasting with varying patterns.
Additionally, STFT can help filter out noise by isolating certain frequency components.

Low-frequency signals capture long-term trends rather than short-term fluctuations.
The choice of window size and overlap significantly affects results, and often requires experimentation.
The STFT can be computationally intensive, especially with large datasets and small window sizes.
While shorter windows improve time resolution, they may reduce frequency resolution, and vice versa.
Furthermore, the extracted features can sometimes lack interpretability, and STFT may struggle to capture complex non-linear relationships in the data without advanced ML techniques.
    \subsubsection{Reinforcement Learning (RL) for TS}
RL
offers a unique approach to TS forecasting, framing the problem as a sequential decision-making task. Instead of directly predicting future values, an RL agent learns a policy to optimize a reward function that reflects the accuracy of its predictions or some other desired objective.
An autonomous agent analyzes its environment to gather information about its current state and take appropriate actions. In contrast, the environment offers a reward signal, which can be either positive or negative. The agent aims to maximize the expected cumulative reward signal throughout the interaction, \cite{ernst2024introduction}.
An agent's policy dictates how it decides to act based on the information available to it. A policy can be deterministic or stochastic, and it may be fixed or contingent upon historical data.

RL agents can adapt to changing patterns in TS data by learning from their past actions and rewards. This capability is especially beneficial in non-stationary environments, where the underlying dynamics of the TS evolve over time.
In principle, RL manages complex and non-linear dependencies within TS data, as it learns to map observations to actions that optimize the reward function.
RL allows incorporating external information or side information into the forecasting process. This can involve incorporating macroeconomic indicators, weather data, or other relevant factors that affect the target TS.
RL can be designed to optimize multiple objectives simultaneously, such as prediction accuracy and risk management.
Training RL agents can be computationally expensive, especially for complex problems and large datasets. The process often requires extensive simulation and exploration of the state-action space.
Carefully designing the reward function is critical. A poorly designed reward function can lead to suboptimal or unexpected agent behaviour. This requires deep domain expertise.
RL algorithms typically require a large amount of training data to learn effectively.
Understanding why an RL agent makes a particular prediction can be challenging, making it a black-box approach in many cases.
Balancing exploration of new actions with exploitation of known good actions is crucial in RL, and finding this balance can be difficult.

    \subsubsection{Graph Neural Network (GNN) for TS}
GNNs
offer a novel approach to TS forecasting by representing the TS data as a graph. This allows for capturing relationships between different time points or incorporating external information represented as nodes and edges in the graph.
GNNs can effectively capture complex, non-linear relationships between different parts of the TS or between the TS and external factors represented as nodes in the graph. This is particularly advantageous when there are dependencies that are not easily captured by sequential models.

External information, including sensor readings, geographic data, and social media sentiment, can be directly integrated into the graph structure to enhance the predictive model.
GNNs can manage TS data with irregular sampling more effectively than some other methods, as the graph structure can represent temporal relationships regardless of the specific sampling intervals.
GNNs are inherently well-suited for managing multivariate TS, where multiple TS are interconnected. This is because graphs can effectively represent the relationships among different series.
Spatial information can be incorporated into the GNN to enhance its efficiency, as demonstrated in numerous studies, including \cite{liang2022survey} and \cite{bui2022spatial}

Defining the appropriate graph structure is crucial. The selection of nodes and edges significantly impacts performance. There is no universally optimal approach; it often requires domain expertise.
Training GNNs can be computationally intensive, particularly when dealing with large graphs and intricate architectures.
Understanding the learned representations within a GNN can be challenging, much like other DL models.
GNNs may require more data than simpler models, especially to effectively learn the graph structure and relationships.
Compared to traditional TS models or RNNs and CNNs, the application of GNNs to TS forecasting is a relatively new field.

\section{Outlier Handling Algorithms}\label{S5}
There are two ways to interpret outliers. First, outliers might
result from measurement errors, such as operator mistakes or equipment
malfunction. In this case, the source of the error needs to be identified and
corrected. However, if we're confident in the accuracy of our data collection
and still observe outliers, we need to consider broader implications.
For instance, unusual weather events like unprecedented floods, heatwaves, or
cold snaps in meteorological data require careful consideration and should
inform future planning and policy decisions.

Outlier predictions encompass various aspects in TS and ML algorithms.
Two primary aspects include the prediction of outlier happening and the
prediction based on outliers. The first aspect can be framed as either a
classification or regression problem, which involves predicting whether
outliers occur in the future and, if they do, estimating their
potential values. The second aspect is specifically a regression problem that
aims to forecast future outcomes using data that contains outliers.
Different algorithms are employed for classification and regression tasks. In
this context, we concentrate on forecasting future outcomes using data that
contains outliers. Let's now explore some ML algorithms suitable for
analyzing this type of TS data.

\cite{lin2020anomaly} introduced a modified version of LSTM
networks combined with a variational autoencoder for outlier detection in TS
data. This algorithm demonstrated the capability to identify anomalies across
multiple time scales. \cite{tang2023gru} proposed GRU networks to address the
challenges associated with multivariate TS data with outliers. They identified GRU
as a viable solution for the problems of vanishing and exploding gradients,
which often arise when using a high number of Hidden Layers (HL)s to learn the
outliers present in the data. \cite{lin2022time} examined LSTM and
GRU networks for detecting outliers in TS data. The findings indicated that GRU
outperformed LSTM due to its higher accuracy and simpler structure, which
requires fewer learning parameters. Bidirectional algorithms have been
studied for outlier detection, and new versions specifically designed for
this type of TS data have been introduced, such as those in
\cite{ullah2021cnn} and \cite{zhang2021air}.

\begin{sidewaystable}
 \caption{Evaluation metrics for test prediction sets. The term Num in the heading row indicates the number of Epochs ($E$), Trees ($T$), Chains ($C$), Iterations per chain ($I$), Levels ($L$), and Burn-in iterations ($B$) for the corresponding algorithms.}\label{EvaluationMetricsForAll}
 \centering
 \small
 \begin{tabular}{cccc}
 \toprule
 \multirow{2}{*}{Method}& Sunspot with mean $81.78$ & CPU with mean $89.79$ & CO with mean $2.15$ \\  \cmidrule{2-4}
  & (Num, MAE, MSE, RMSE) & (Num, MAE, MSE, RMSE) & (Num, MAE, MSE, RMSE) \\ \midrule
 ARIMA & (-, 48.31, 3822.78, 61.83) & (-, 94.57, 9018.92, 94.97) & (-,4.69, 24.63, 4.96) \\
 Elman & $ ( E:7, 248.41,63434.72 , 251.86) $ & $ ( E:10,4.21, 19.14, 4.38) $ & $ (E: 10, 1.98,4.59 , 2.14 ) $ \\
 LSTM & $ (E: 81, 47.40,3417.84 ,58.46 ) $ & $ (E:220, 0.92,1.47 ,1.21 ) $ & $ (E: 100,1.30 ,2.83 ,1.68 ) $ \\
 GRU & $ (E:10 ,69.54 ,6264.28 ,79.15 ) $ & $ (E:10 ,1.02 ,1.75 ,1.32 ) $ & $ (E: 10,0.75 , 1.19,1.09 ) $ \\
 Bidirectional RNN & $ (E:10 , 51.99,3462.63 ,58.84 ) $ & $ (E:10 , 0.92,1.44 , 1.20) $ & $ (E: 10,1.50 ,3.34 , 1.83) $ \\
 Deep RNN & $ (E:10 ,194.84 ,43647.04 ,208.92 ) $ & $ (E:10 , 4.45,21.26 ,4.61 ) $ & $ (E:10 ,1.50 ,2.78 ,1.67 ) $ \\
 DeepAR & $ (E:100 , 15.88, 417.37,20.43 ) $ & $ (E:100 , 1.47,3.50 ,1.87 ) $ & $ (E:100 , 0.90 ,1.55 ,1.25 ) $ \\
 RF & $ (T:100  ,9.84 ,220.04 ,14.83  ) $ & $ (T:100  ,0.98 , 1.53, 1.24 ) $ & $ (T:100  , 0.35, 0.28,0.53 ) $ \\
 XGBoost & $ (T: 100, 10.61,264.13 ,16.25 ) $ & $ (T:100 , 1.01,1.61 , 1.27) $ & $ (T:100 ,0.36 , 0.29,0.54 ) $ \\
 LightGBM & $ (T: 1000, 6.21,73.26 , 8.56) $ & $ (T:1000 , 0.18, 0.07, 0.27) $ & $ (T: 1000,0.06 , 0.01, 0.08) $ \\
 Prophet & $ ([C:4, I:30] ,40.54 ,2026.03 ,45.01 ) $ & $ ([C:4 , I:100] , 1.10, 1.86,1.36 ) $ & $ (I: 10000,0.75 ,0.93 , 0.96) $ \\
 WBT & $ ([L:5, T:100] ,11.42 , 252.43,15.89  ) $ & $ ([L:5, T:100] ,0.93 ,1.51 , 1.23) $ & $ ([L:5, T:100] , 0.36 ,  0.27, 0.52) $ \\
 TFT & $ (E:25 , 40.56,2064.00 ,45.43 ) $ & $ (E:15 ,22.78 ,525.74 , 22.93) $ & $ (E: 26, 6.15,  39.16,6.26 ) $ \\
 N-BEATS & $ (E: 100, 10.79,231.46 , 15.21) $ & $ (E: 100, 0.96, 1.56,1.25 ) $ & $ (E: 100,3.04 ,9.25 ,2.93 ) $ \\
 N-HiTS & $ (E:100 ,72.28 , 6852.23, 82.78) $ & $ (E: 100,4.92 ,28.82 ,5.37 ) $ & $ (E:100 ,  3.60,17.62 ,4.20 ) $ \\
 DFFNN & $ (E: 10,104.57 ,12952.18 ,113.81 ) $ & $ (E:30 , 5.12, 28.32,5.32 ) $ & $ (E:30 ,1.21 , 2.12 , 1.46) $ \\
 BNN & $ ([C:2,B:1000, I:2000]  , $ & $ ([C:2,B:1000, I:2000]  , $ & $ ([C:2,B:1000, I:2000]  , $ \\
 & $ 49.72,3144.41 , 56.08 ) $ & $ 4.97 ,  26.18,5.12 ) $ & $ 1.05 ,  1.56,1.25 ) $ \\
 CNN & $ (E: 100,13.56 , 294.84, 17.17) $ & $ (E:100 ,1.01 ,1.64 , 1.28) $ & $ (E: 100,0.56 , 0.88 , 0.94) $ \\
 TCN & $ (E: 30,12.04 ,259.91 , 16.12) $ & $ (E:30 , 4.60,22.61 ,4.75 ) $ & $ (E: 100,0.37 ,0.30 , 0.54) $ \\
 STFT & $ (E:500  ,49.95 ,3376.51 , 58.11 ) $ & $ (E: 500 , 4.73,24.19 ,4.92 ) $ & $ (E: 500 , 1.06 ,1.60 , 1.26) $ \\
 RL & $ (E: 90, 39.77, 2344.65, 48.42) $ & $ (E: 90, 1.37, 3.07, 1.75) $ & $ (E: 90,1.24 ,2.70 ,1.64 ) $ \\
 GNN & $ (E: 100, 12.79, 322.76, 17.97) $ & $ (E:100 ,1.45 ,4.86 ,2.20 ) $ & $ (E: 100,0.50 ,0.46 , 0.68) $ \\
 \bottomrule
 \end{tabular}
\end{sidewaystable}
\begin{figure}[t] 
\begin{subfigure}[b]{1\linewidth}
\lineskip=0pt
\includegraphics[scale=0.6, trim={5pt 5pt 0pt 20pt}, clip]{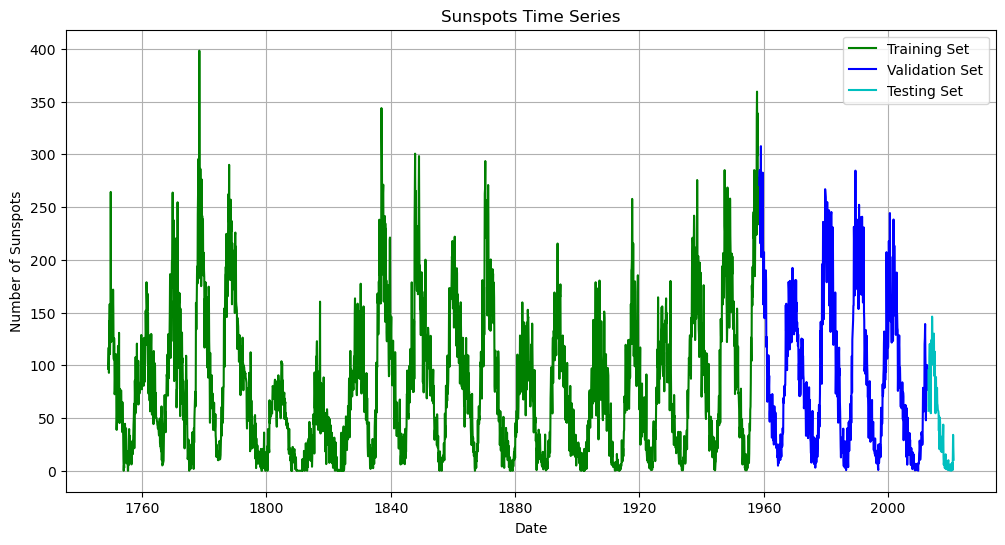}%
\end{subfigure}
\rule{\textwidth}{0.1pt}
\caption{Sunspot TS. The sizes of the training, validation, and testing sets are $2514$, $653$, and $98$, respectively.}\label{SunspotPlot}
\end{figure}
\begin{figure}[b]
\begin{subfigure}[b]{1\linewidth}
\lineskip=0pt
\includegraphics[scale=0.3, trim={5pt 5pt 0pt 20pt}, clip]{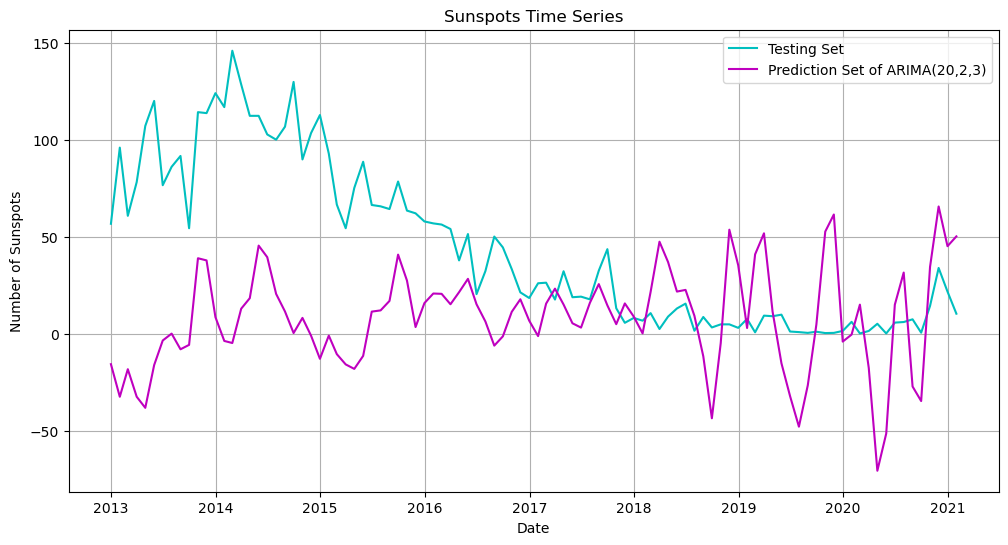}%
\includegraphics[scale=0.3, trim={20pt 5pt 0pt 20pt}, clip]{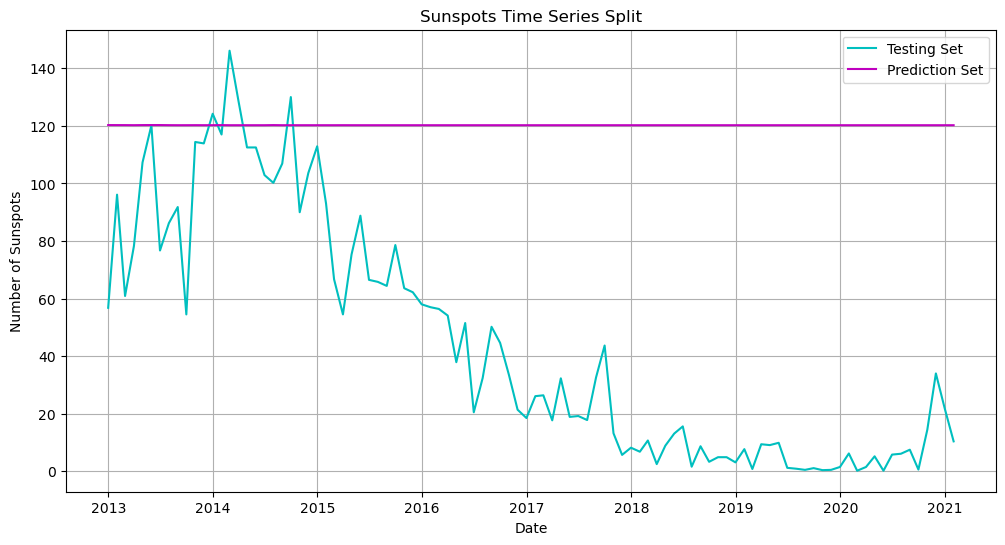}%
\caption{ARIMA and Elman.}
\end{subfigure}
\rule{\textwidth}{0.1pt}
\caption{Continue.}\label{SunResults}
\end{figure}
\begin{figure}[h]
\ContinuedFloat
\begin{subfigure}[b]{1\linewidth}
\lineskip=0pt
\includegraphics[scale=0.3, trim={5pt 5pt 0pt 20pt}, clip]{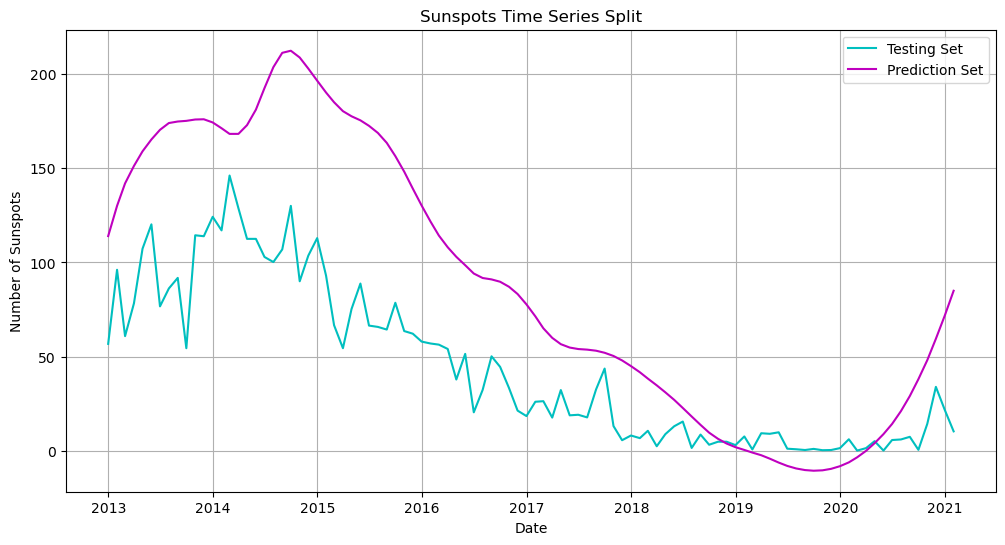}%
\includegraphics[scale=0.3, trim={20pt 5pt 0pt 20pt}, clip]{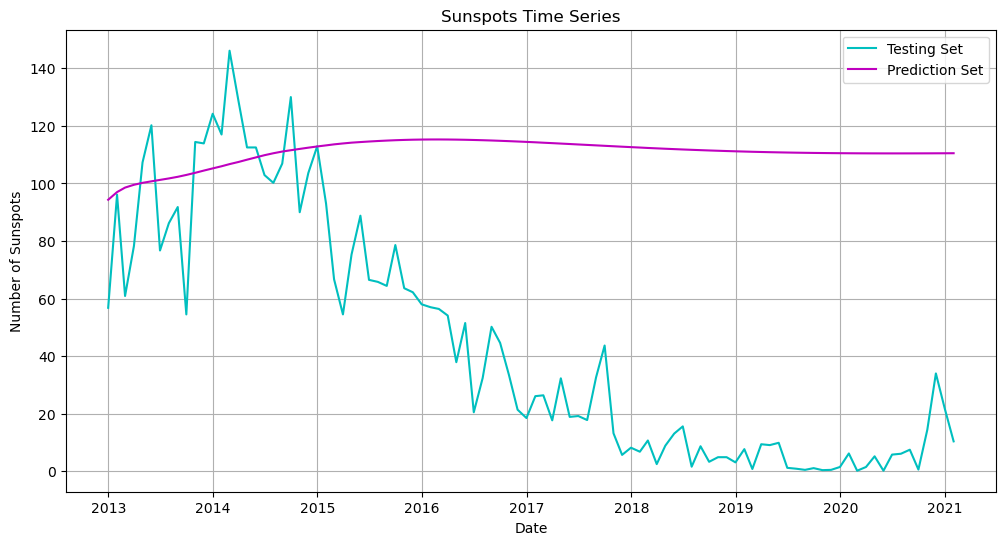}%
\caption{LSTM and GRU.}
\end{subfigure}
\begin{subfigure}[b]{1\linewidth}
\lineskip=0pt
\includegraphics[scale=0.3, trim={5pt 5pt 0pt 20pt}, clip]{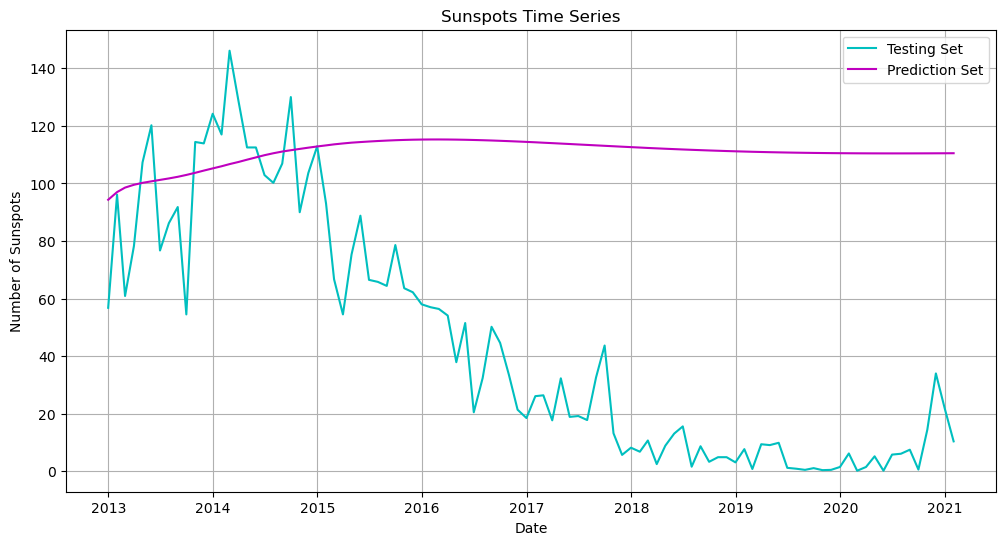}%
\includegraphics[scale=0.3, trim={20pt 5pt 0pt 20pt}, clip]{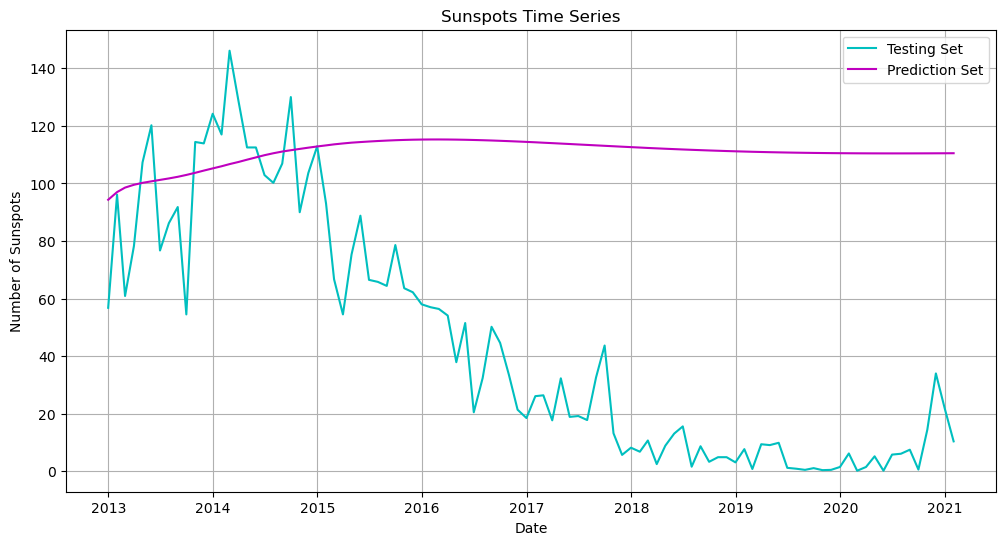}%
\caption{Bidirectional LSTM and Deep Elman.}
\end{subfigure}
\begin{subfigure}[b]{1\linewidth}
\lineskip=0pt
\includegraphics[scale=0.3, trim={5pt 5pt 0pt 20pt}, clip]{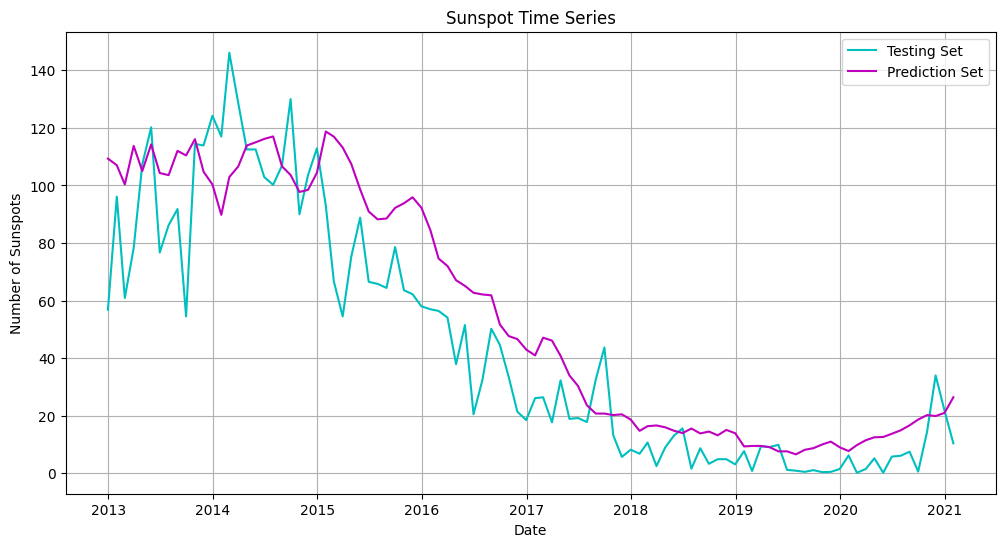}%
\includegraphics[scale=0.3, trim={20pt 5pt 0pt 20pt}, clip]{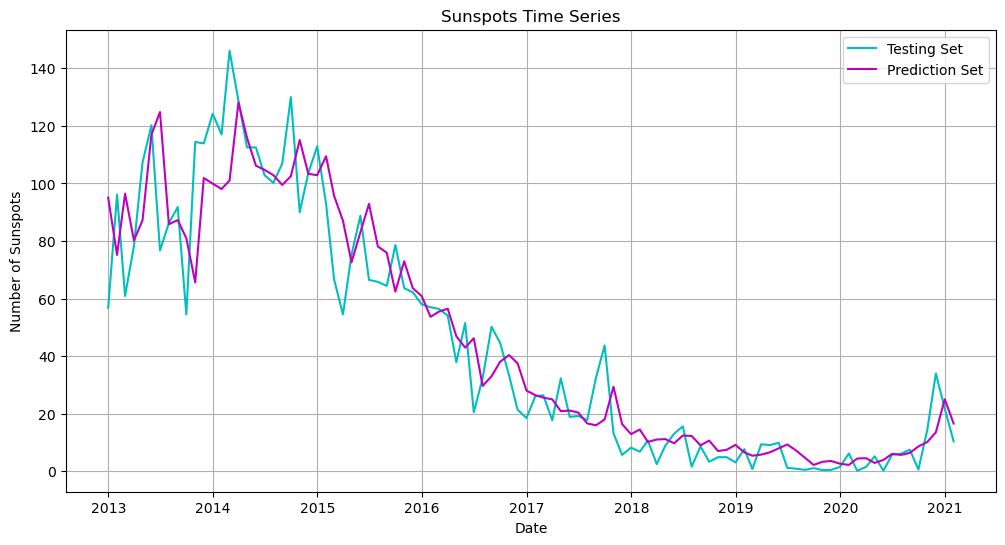}%
\caption{DeepAR and RF.}
\end{subfigure}
\begin{subfigure}[b]{1\linewidth}
\lineskip=0pt
\includegraphics[scale=0.3, trim={20pt 5pt 0pt 20pt}, clip]{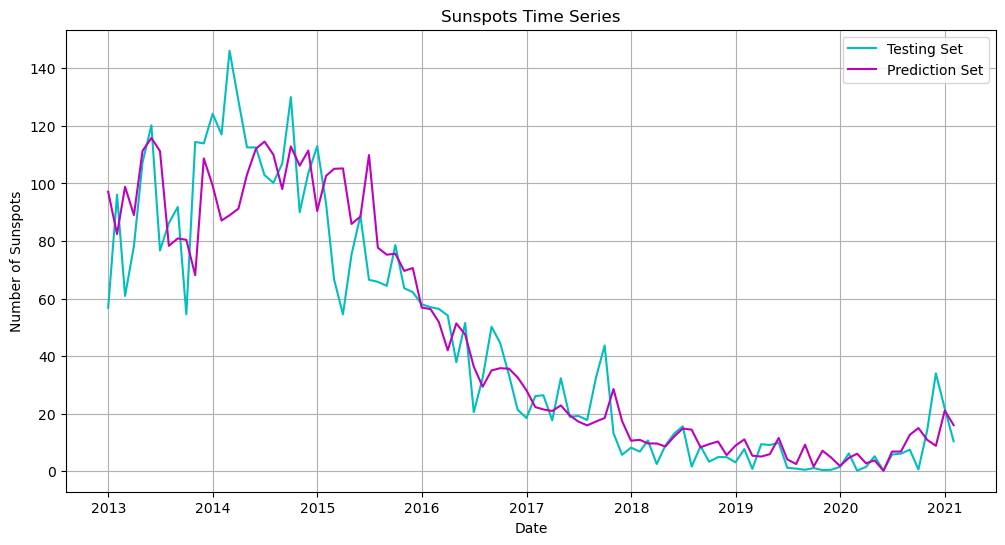}%
\includegraphics[scale=0.3, trim={5pt 5pt 0pt 20pt}, clip]{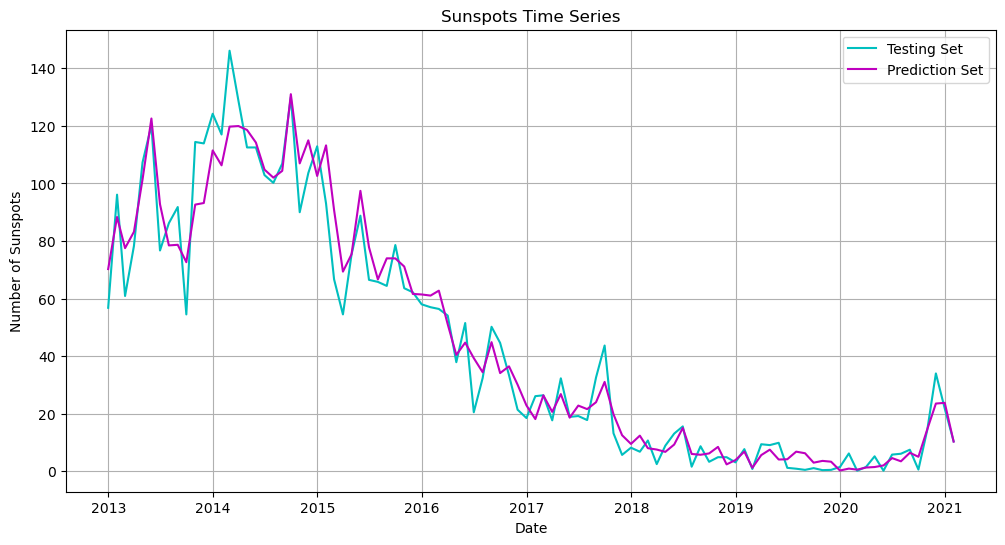}%
\caption{XGBoost and LightGBM.}
\end{subfigure}
\rule{\textwidth}{0.1pt}
\caption{Continue.}\label{SunResults}
\end{figure}
\begin{figure}[h]
\ContinuedFloat
\begin{subfigure}[b]{1\linewidth}
\lineskip=0pt
\includegraphics[scale=0.3, trim={20pt 5pt 0pt 20pt}, clip]{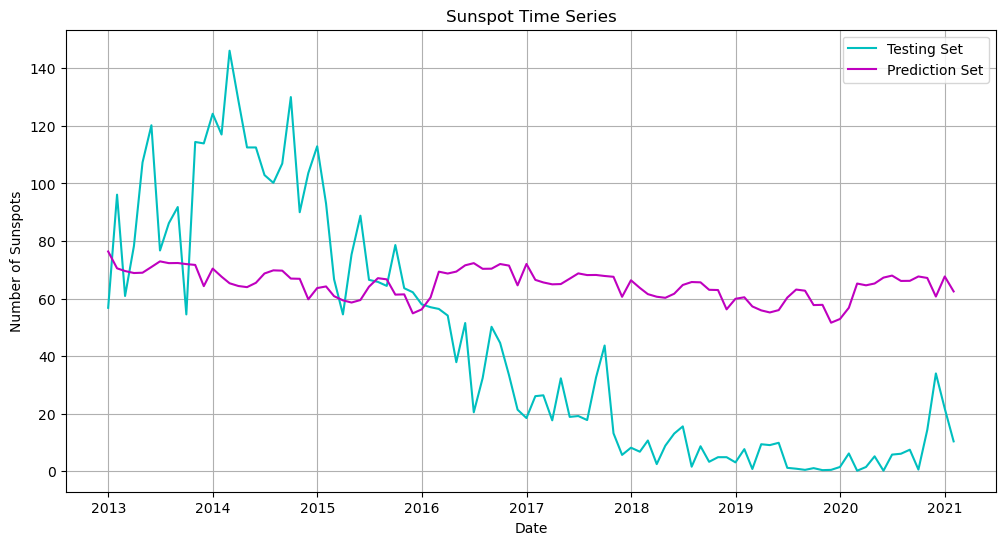}%
\includegraphics[scale=0.3, trim={20pt 5pt 0pt 20pt}, clip]{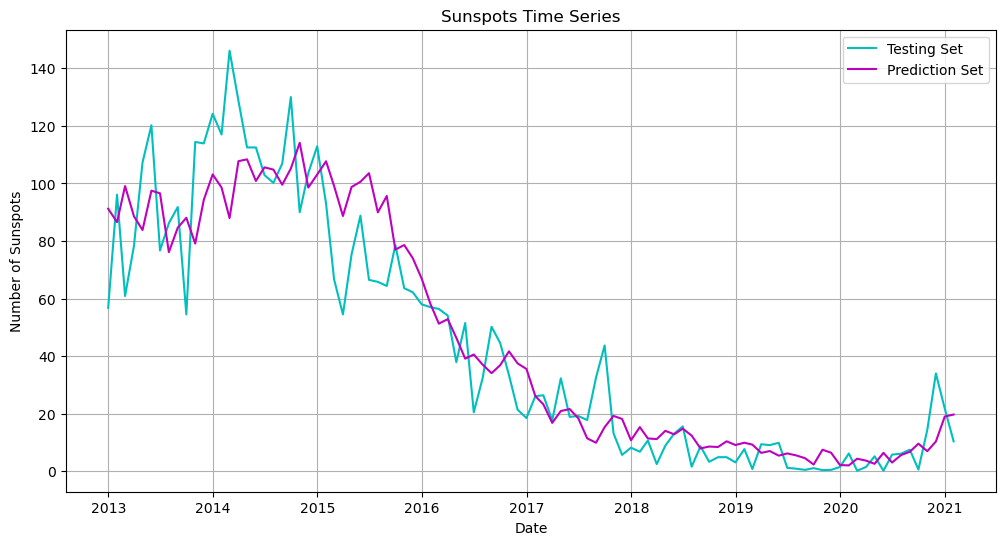}%
\caption{Prophet and WBT.}
\end{subfigure}
\begin{subfigure}[b]{1\linewidth}
\lineskip=0pt
\includegraphics[scale=0.3, trim={5pt 5pt 0pt 20pt}, clip]{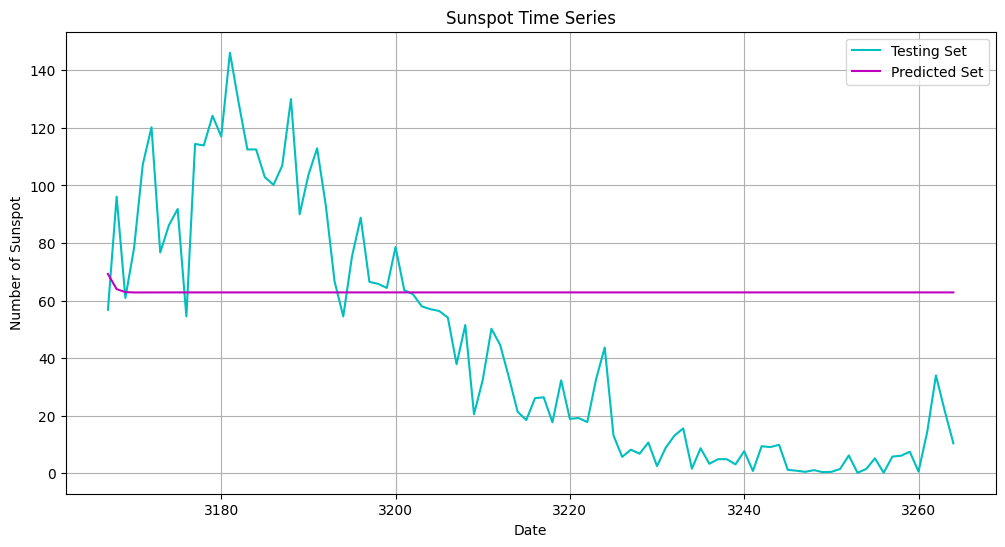}%
\includegraphics[scale=0.3, trim={20pt 5pt 0pt 20pt}, clip]{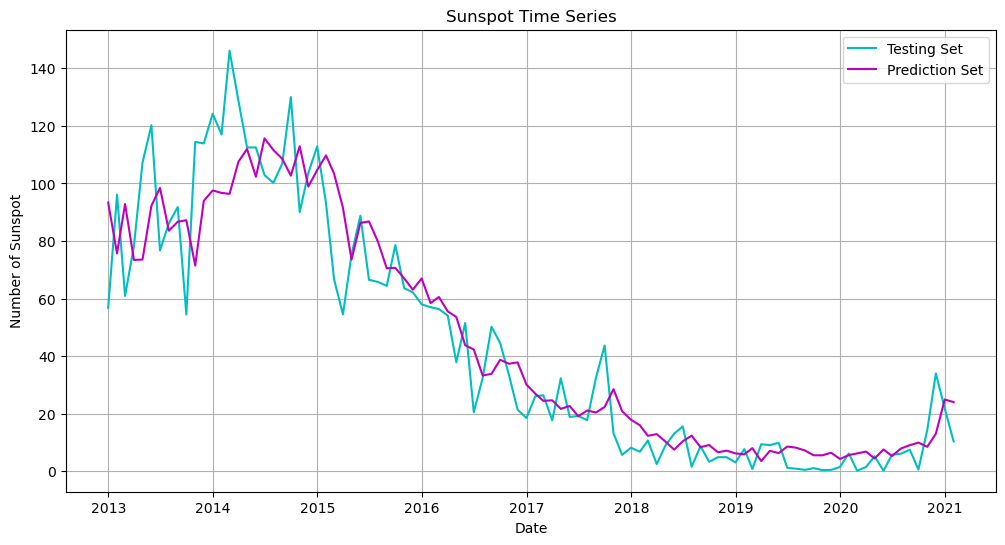}%
\caption{TFT and N-BEATS.}
\end{subfigure}
\begin{subfigure}[b]{1\linewidth}
\lineskip=0pt
\includegraphics[scale=0.3, trim={5pt 5pt 0pt 20pt}, clip]{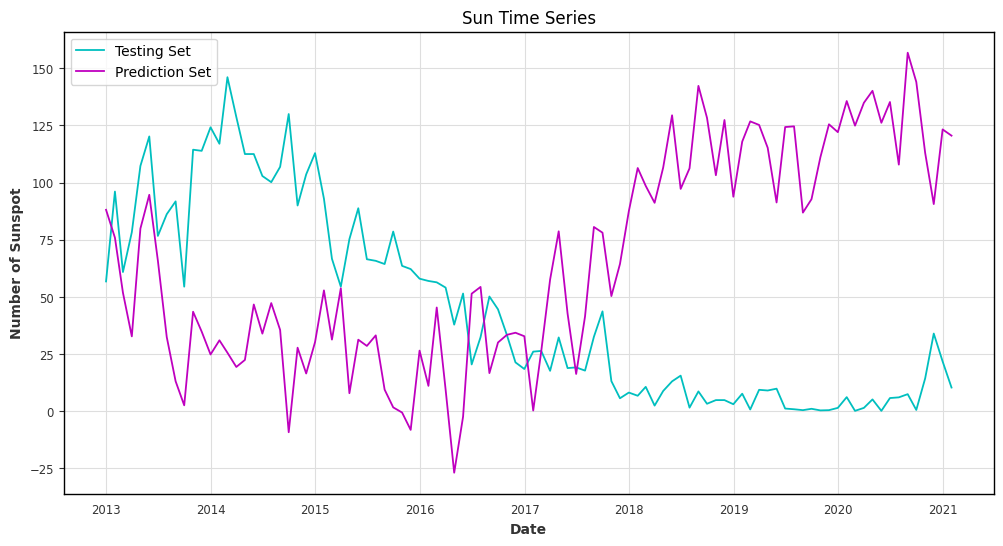}%
\includegraphics[scale=0.3, trim={20pt 5pt 0pt 20pt}, clip]{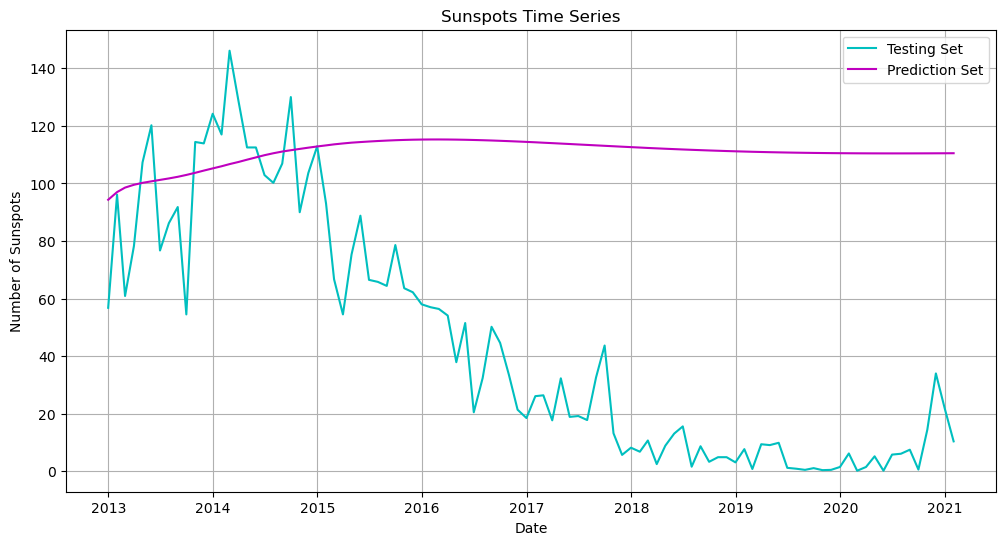}%
\caption{N-HiTS and DFFNN}
\end{subfigure}
\begin{subfigure}[b]{1\linewidth}
\lineskip=0pt
\includegraphics[scale=0.3, trim={5pt 5pt 0pt 20pt}, clip]{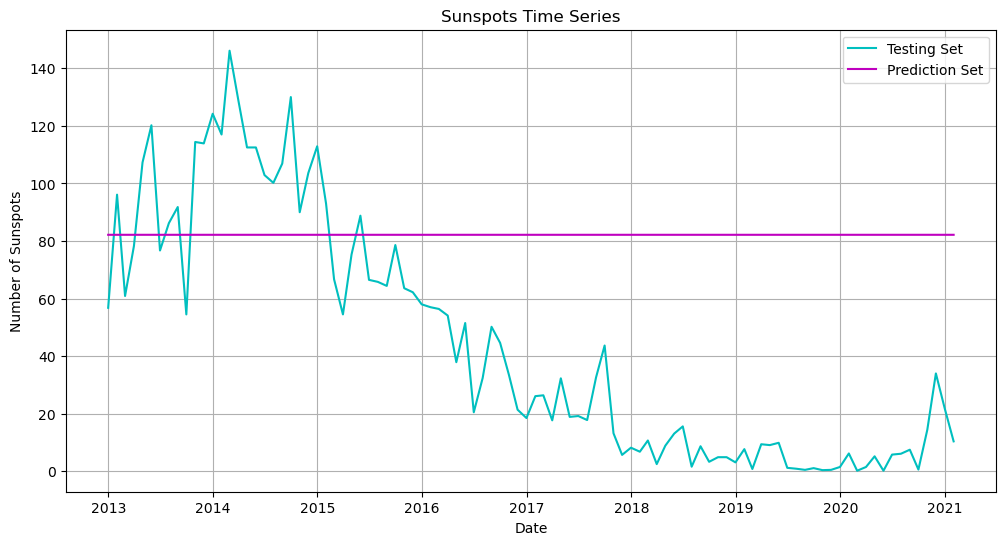}%
\includegraphics[scale=0.3, trim={5pt 5pt 0pt 20pt}, clip]{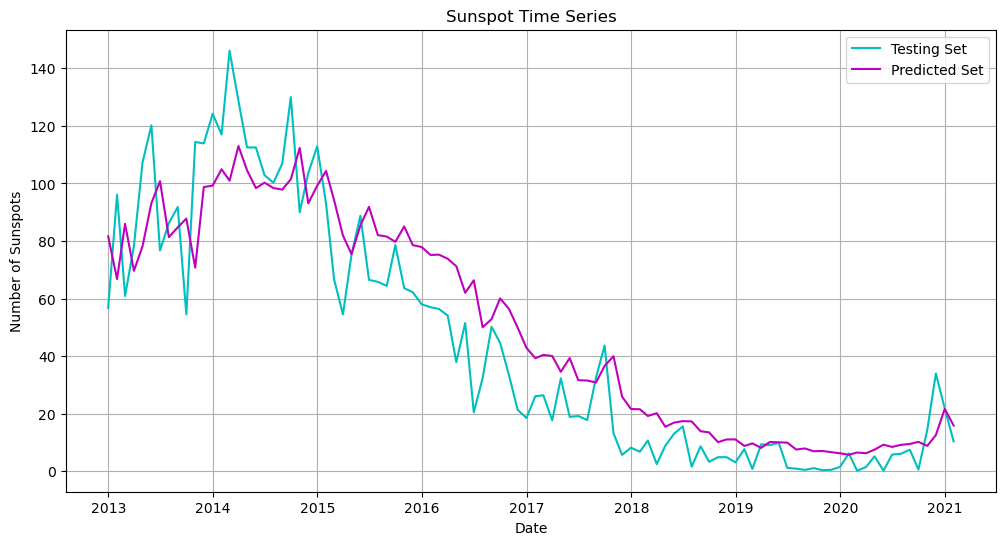}%
\caption{BNN and CNN.}
\end{subfigure}
\rule{\textwidth}{0.1pt}
\caption{Continue.}\label{SunResults}
\end{figure}
\clearpage
\begin{figure}[t]
\ContinuedFloat
\begin{subfigure}[b]{1\linewidth}
\lineskip=0pt
\includegraphics[scale=0.3, trim={20pt 5pt 0pt 20pt}, clip]{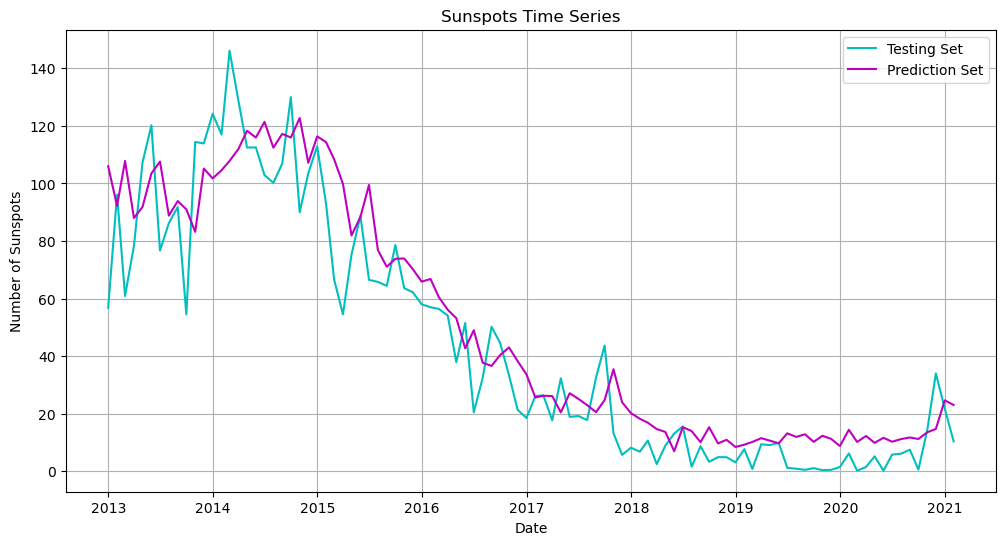}%
\includegraphics[scale=0.3, trim={20pt 5pt 0pt 20pt}, clip]{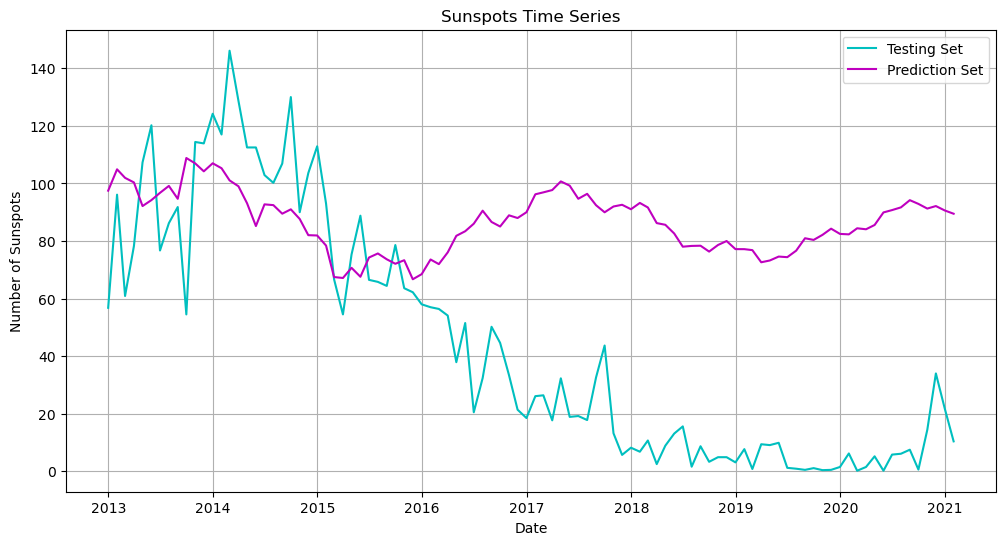}%
\caption{TCN and STFT.}
\end{subfigure}
\begin{subfigure}[b]{1\linewidth}
\lineskip=0pt
\includegraphics[scale=0.3, trim={5pt 5pt 0pt 20pt}, clip]{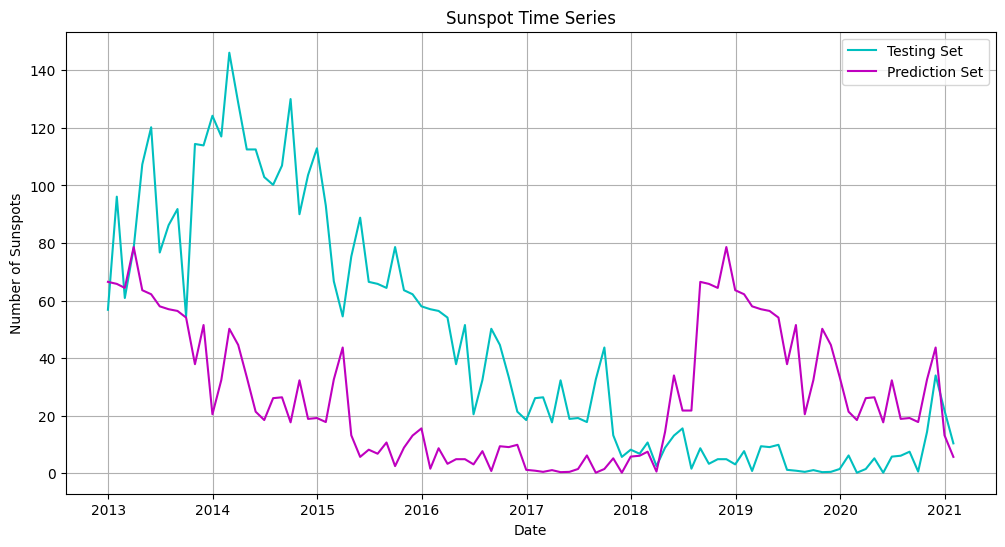}%
\includegraphics[scale=0.26, trim={20pt 5pt 0pt 20pt}, clip]{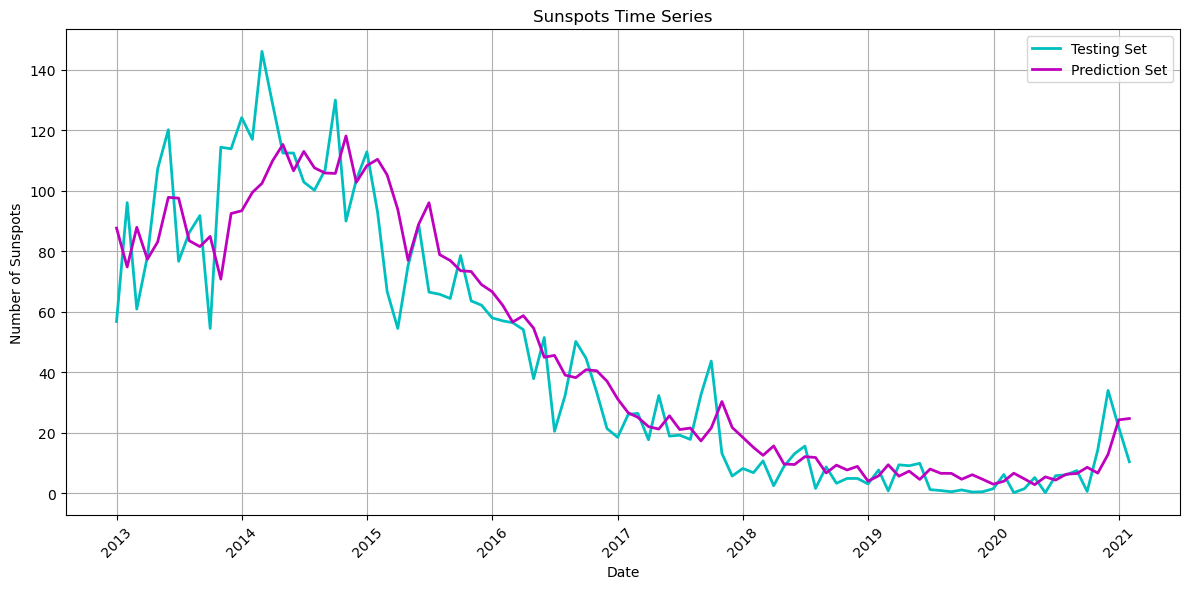}%
\caption{RL and GNN.}
\end{subfigure}
\rule{\textwidth}{0.1pt}
\caption{Prediction results for Sunspot TS.}\label{SunResults}
\end{figure}

Other algorithms renowned for outlier detection include tree-based ensemble algorithms.
\cite{zhang2022anomaly} applied XGBoost and LSTM-stacked denoising
autoencoders to effectively predict outliers in wind patterns in
northeastern China. The study by \cite{hartanto2023stock} addressed the
approximation of stock price TS using LightGBM and compared the results with
XGBoost, AdaBoost, and CatBoost. Their experiment demonstrated the superior
capabilities of LightGBM in forecasting stock price data, even when it
contained outliers.

\cite{zhang2022prediction} introduced an enhanced version of the N-BEATS
network that integrates seasonality with support vector machines to diagnose
outliers in TS data related to sewage treatment. \cite{falchi2023deep}
involved training a TFT model to predict the vibrational characteristics of
the aged Guinigi Tower's experimental frequencies, located in Lucca, Italy,
along with various environmental parameters to monitor the structural health
of the tower. The primary objective of the TFT model was to identify
potential damages caused by the Viareggio earthquake that occurred in 2022.
The implemented TFT model demonstrated sensitivity to outliers and provided
accurate predictions.

\cite{dong2021network} proposed a semi-supervised Double Deep Q-Network
method as a variant of RL for outlier detection, addressing the limitations
of traditional methods when dealing with high-dimensional data and costly
labelling processes. Their experimental results demonstrated the
effectiveness of the RL approach in comparison to existing ML models.

This paper tests the algorithms described above on a TS containing outliers. We then analyze and interpret their performance and effectiveness.
\section{Missing data Handling Algorithms}\label{S6}
Missing data is common in any kind of datasets, particularly when collected over an extended period. Therefore, missing data is a frequent occurrence in TS analysis. Addressing this issue in ML can be categorized into two approaches: some algorithms can directly handle datasets with missing values, while others necessitate the imputation of these missing values using appropriate methods, such as the mean, median, or interpolation.
DeepAR and Prophet can handle data with missing values, while the other algorithms discussed in this paper require the imputation of appropriate values to address these gaps.
There are many research papers on handling this type of data, such as \cite{qin2023imputegan}, where they introduced a model to handle missing data for multivariate TS using generative adversarial networks, incorporating an iterative strategy and gradient optimization. They tested the method by implementing it on three large-scale datasets. As a result, the generative algorithm excelled in accuracy compared to traditional methods.
\section{Real-World Examples}\label{S7}
In this section, we examine three distinct datasets: the average number of sunspots over centuries, CPU usage with outliers, and CO concentration with missing data. Our objective is to compare the performance of the presented ML algorithms in these examples.
For the evaluation criteria, we utilize Mean Absolute Error (MAE), Mean Squared Error (MSE), and Root Mean Squared Error (RMSE). The evaluation values are presented in Table \ref{EvaluationMetricsForAll}, providing an overview of the performance and allowing for a comparison of the errors with the dataset means.

For the ARIMA model, we make the data stationary and normalized before implementing the forecasting process.
For ML algorithms, we disregard the stationary process.
Additionally, we overlook the normalization process for DeepAR, tree-based ensemble methods, WBT, Prophet, N-HITS, BNN, RL, and GNN. We scale the data for the remaining algorithms to achieve better results with lower computational costs.

\begin{figure}[t] 
\begin{subfigure}[b]{1\linewidth}
\lineskip=0pt
\includegraphics[scale=0.6, trim={5pt 5pt 0pt 20pt}, clip]{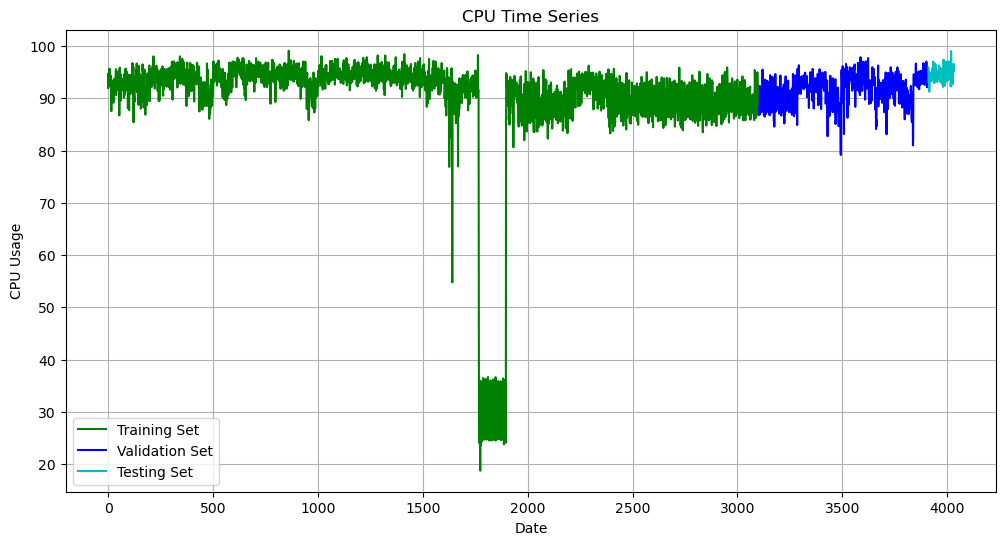}%
\end{subfigure}
\rule{\textwidth}{0.1pt}
\caption{CPU TS. The sizes of the training, validation, and testing sets are $3104$, $807$, and $121$, respectively.}\label{CPUPlot}
\end{figure}
\begin{figure}[b]
\begin{subfigure}[b]{1\linewidth}
\lineskip=0pt
\includegraphics[scale=0.3, trim={5pt 5pt 0pt 20pt}, clip]{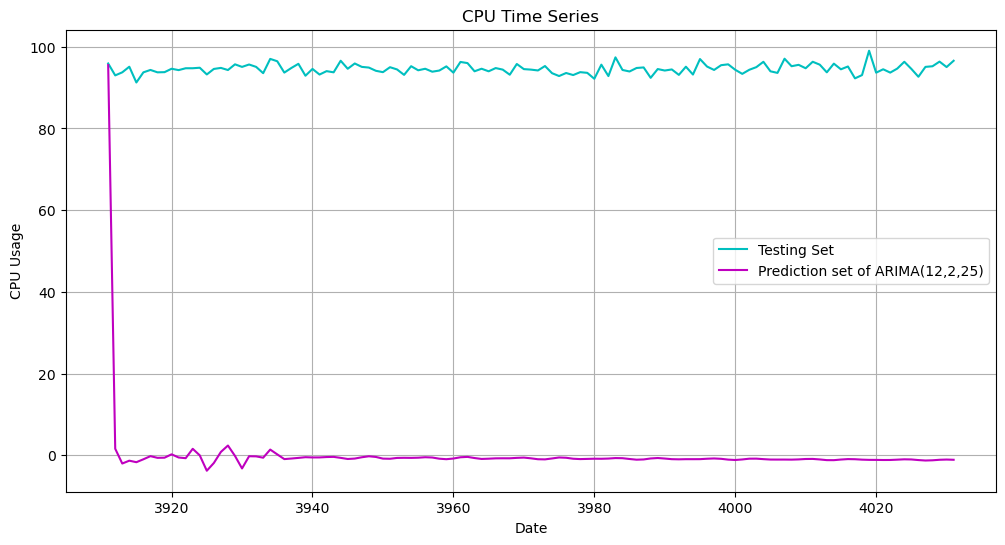}%
\includegraphics[scale=0.3, trim={20pt 5pt 0pt 20pt}, clip]{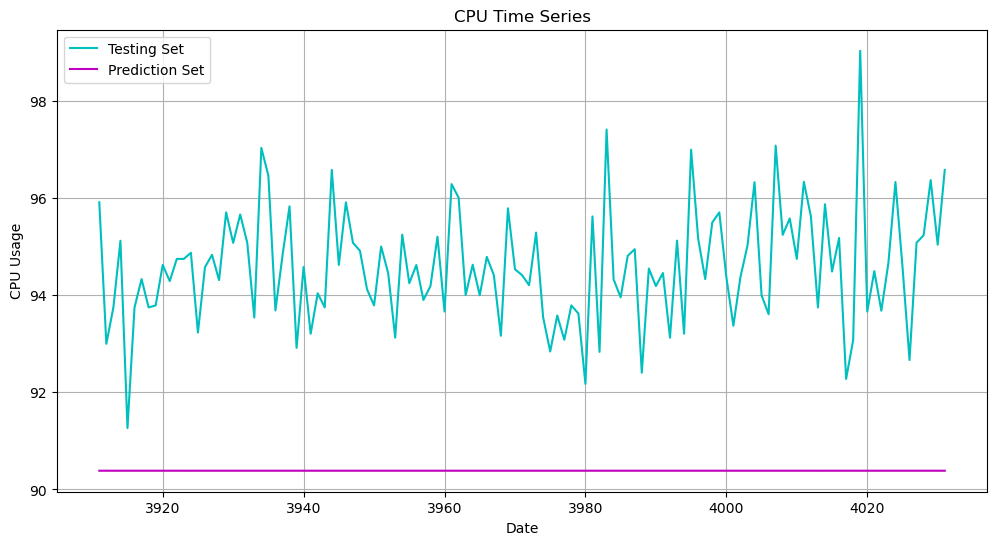}%
\caption{ARIMA and Elman.}
\end{subfigure}
\rule{\textwidth}{0.1pt}
\caption{Continue.}\label{CPUResults}
\end{figure}
\begin{figure}[b]
\ContinuedFloat
\begin{subfigure}[b]{1\linewidth}
\lineskip=0pt
\includegraphics[scale=0.3, trim={5pt 5pt 0pt 20pt}, clip]{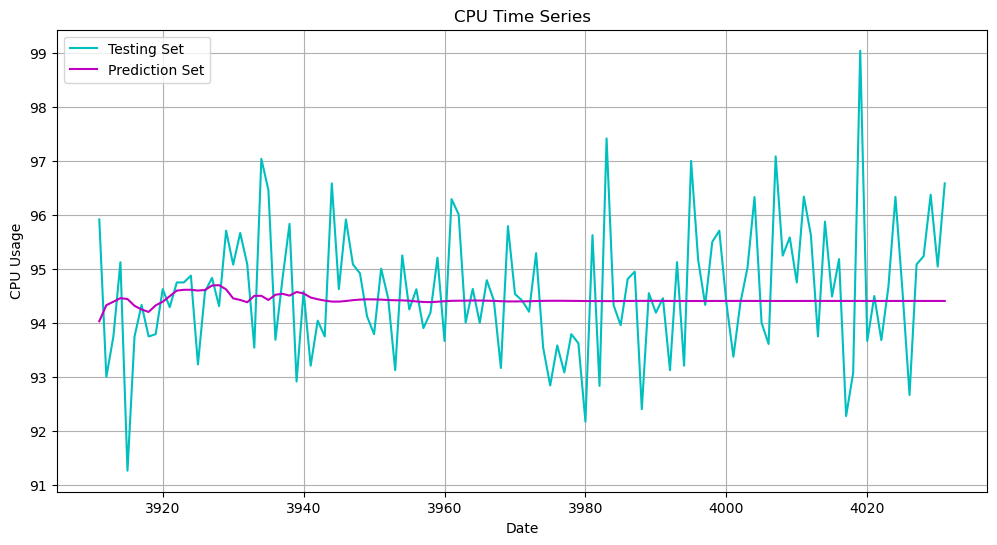}%
\includegraphics[scale=0.3, trim={20pt 5pt 0pt 20pt}, clip]{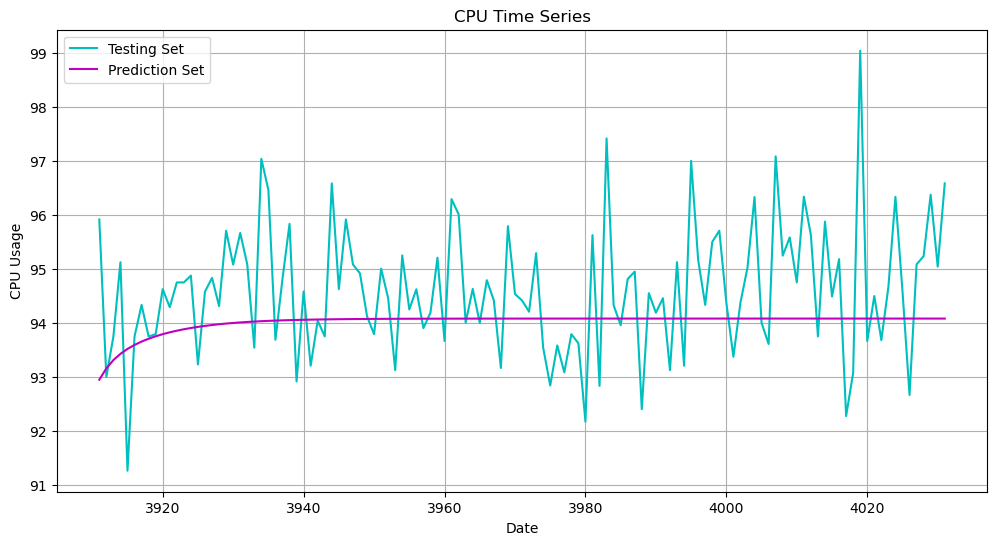}%
\caption{LSTM and GRU.}
\end{subfigure}
\begin{subfigure}[b]{1\linewidth}
\lineskip=0pt
\includegraphics[scale=0.3, trim={5pt 5pt 0pt 20pt}, clip]{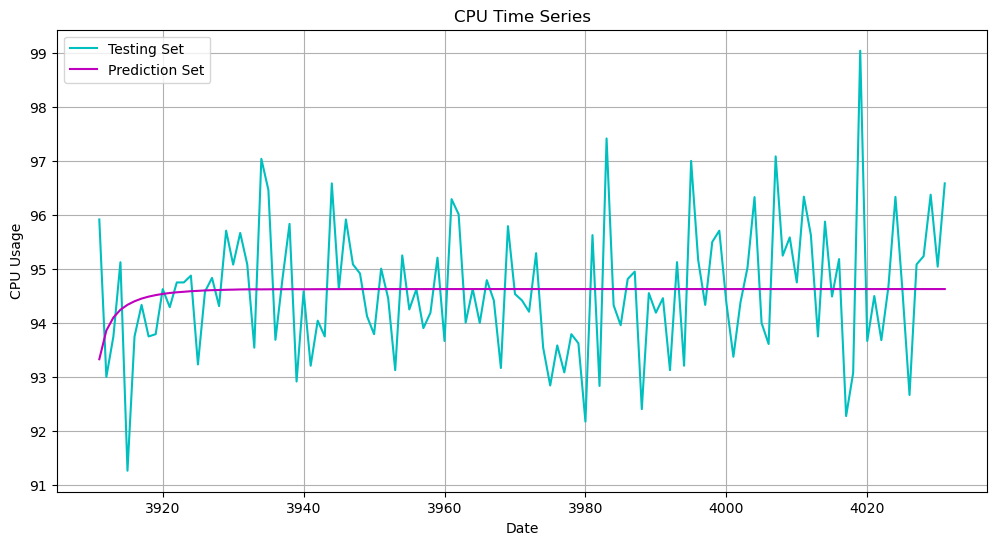}%
\includegraphics[scale=0.3, trim={20pt 5pt 0pt 20pt}, clip]{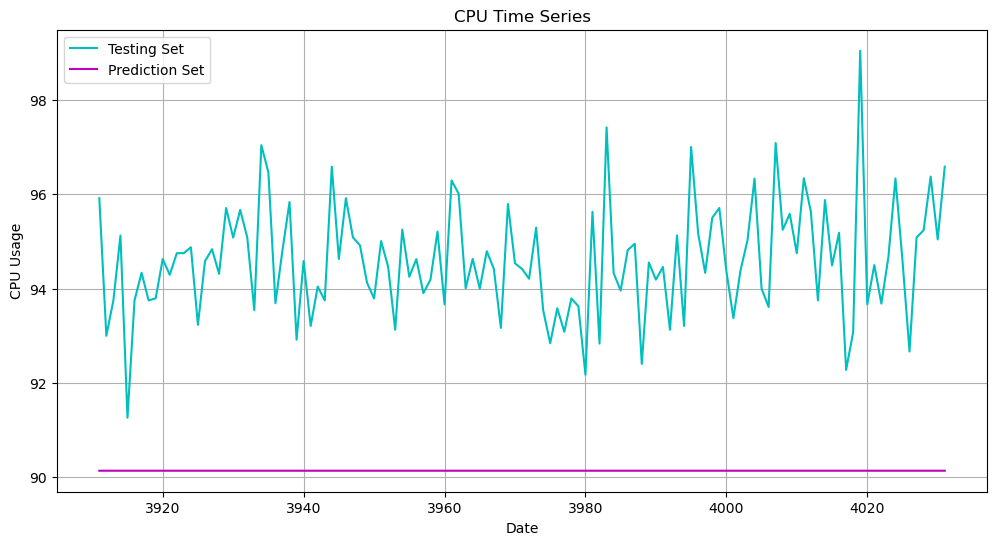}%
\caption{Bidirectional LSTM and Deep Elman.}
\end{subfigure}
\begin{subfigure}[b]{1\linewidth}
\lineskip=0pt
\includegraphics[scale=0.3, trim={5pt 5pt 0pt 20pt}, clip]{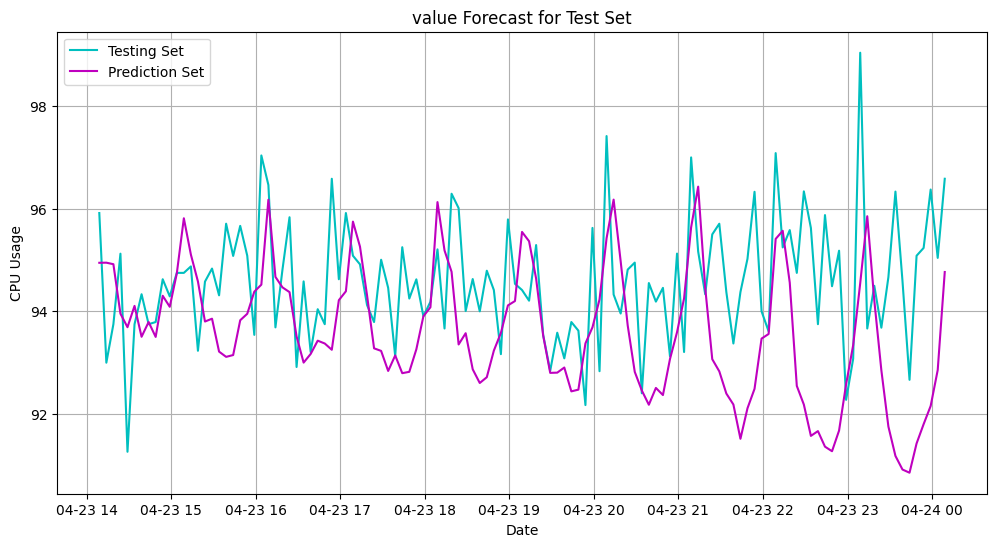}%
\includegraphics[scale=0.3, trim={20pt 5pt 0pt 20pt}, clip]{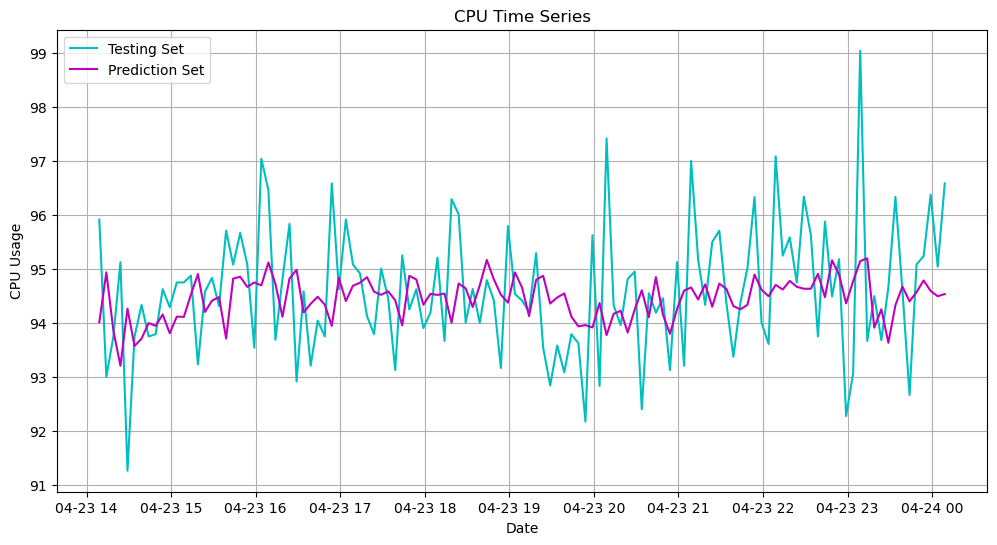}%
\caption{DeepAR and RF.}
\end{subfigure}
\begin{subfigure}[b]{1\linewidth}
\lineskip=0pt
\includegraphics[scale=0.3, trim={20pt 5pt 0pt 20pt}, clip]{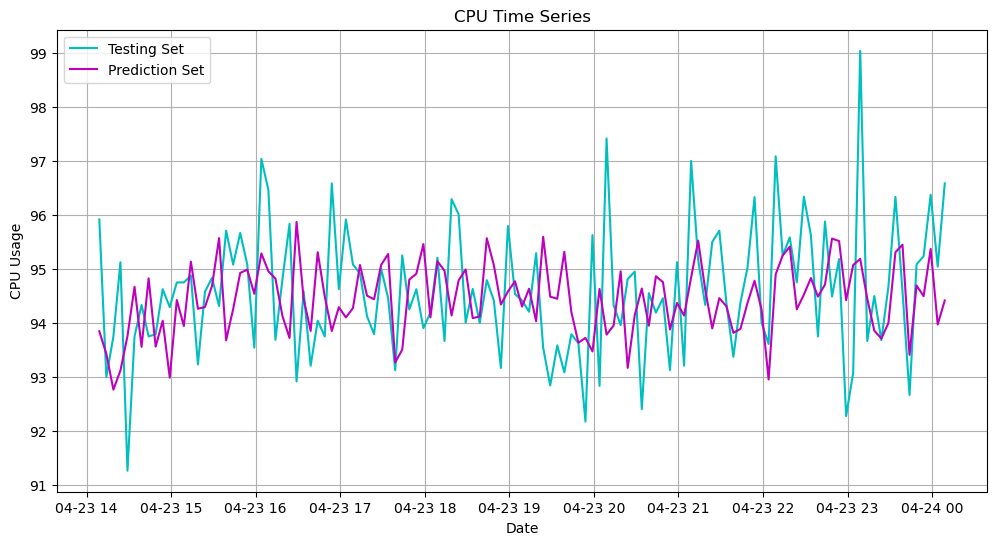}%
\includegraphics[scale=0.3, trim={5pt 5pt 0pt 20pt}, clip]{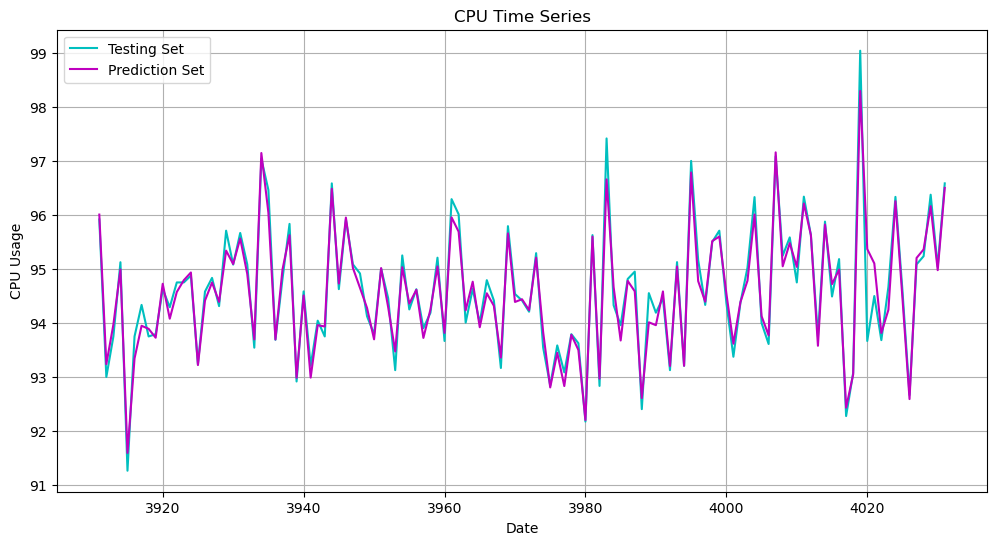}%
\caption{XGBoost and LightGBM.}
\end{subfigure}
\rule{\textwidth}{0.1pt}
\caption{Continue.}\label{CPUResults}
\end{figure}
\begin{figure}[b]
\ContinuedFloat
\begin{subfigure}[b]{1\linewidth}
\lineskip=0pt
\includegraphics[scale=0.3, trim={20pt 5pt 0pt 20pt}, clip]{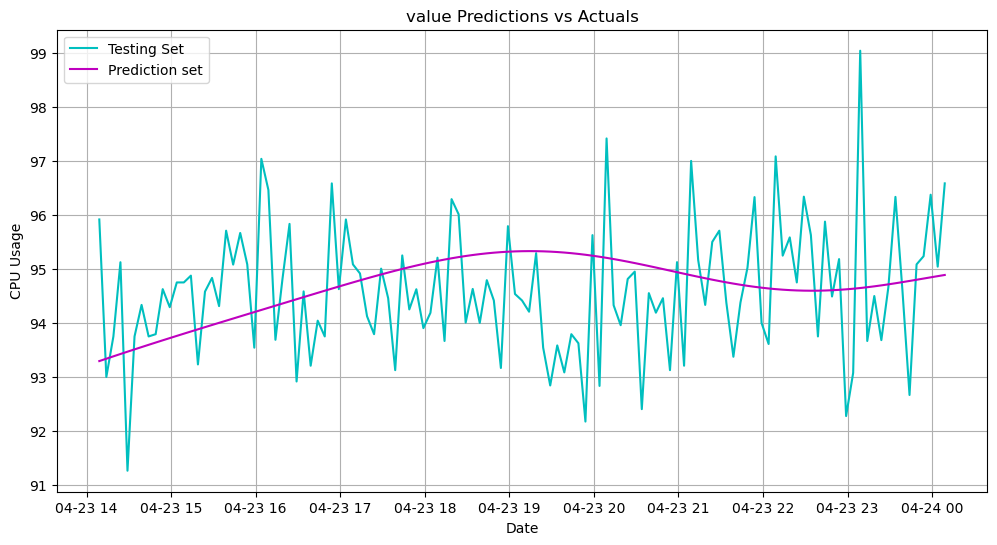}%
\includegraphics[scale=0.3, trim={20pt 5pt 0pt 20pt}, clip]{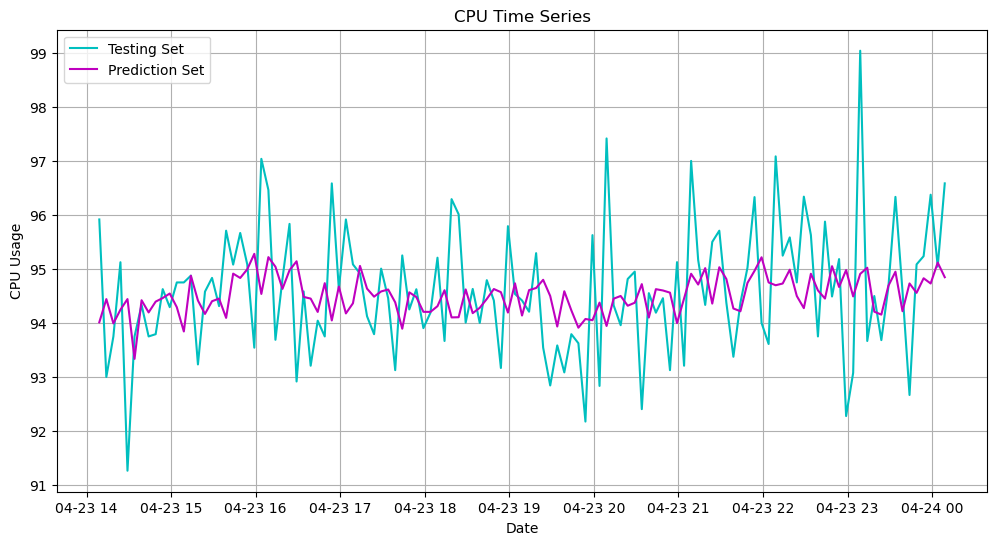}%
\caption{Prophet and WBT.}
\end{subfigure}
\begin{subfigure}[b]{1\linewidth}
\lineskip=0pt
\includegraphics[scale=0.3, trim={5pt 5pt 0pt 20pt}, clip]{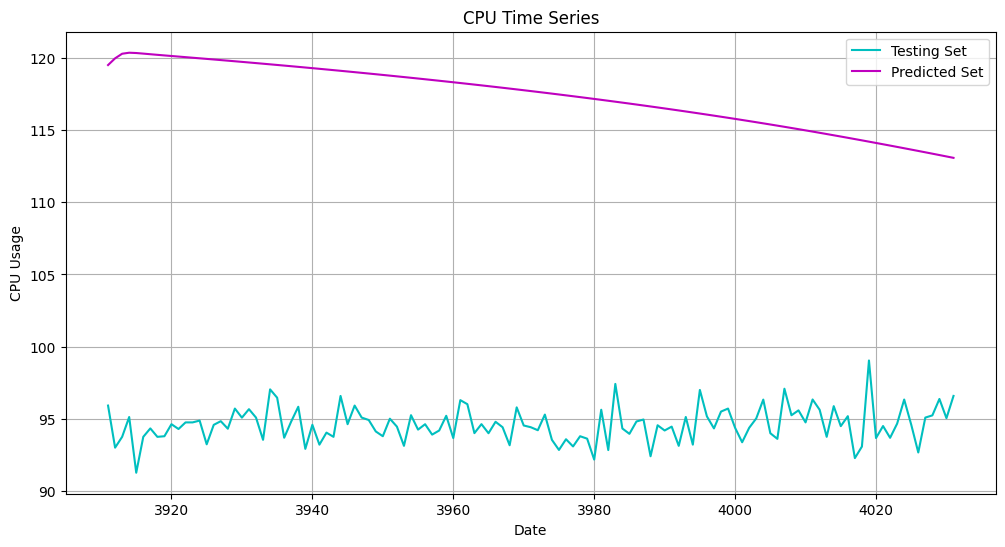}%
\includegraphics[scale=0.3, trim={20pt 5pt 0pt 20pt}, clip]{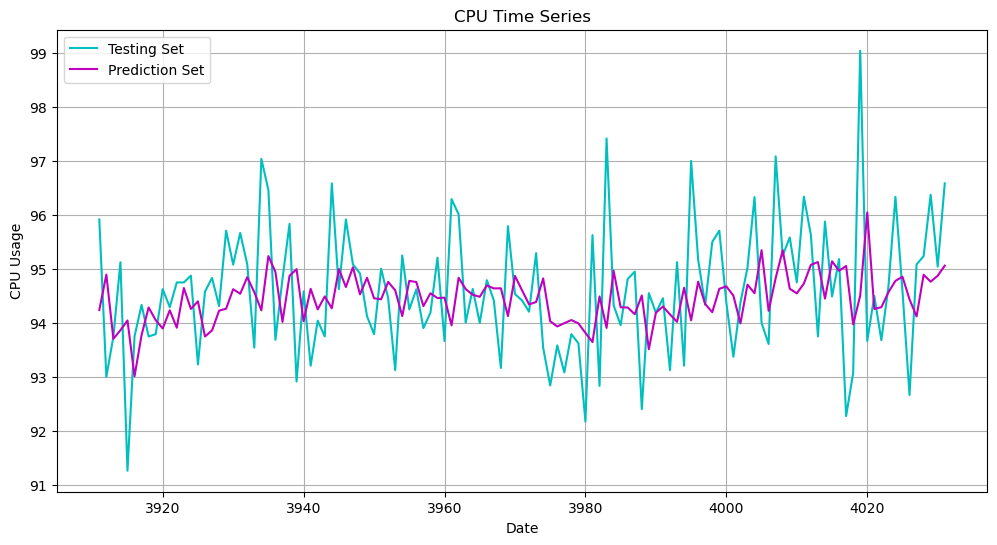}%
\caption{TFT and N-BEATS.}
\end{subfigure}
\begin{subfigure}[b]{1\linewidth}
\lineskip=0pt
\includegraphics[scale=0.3, trim={5pt 5pt 0pt 20pt}, clip]{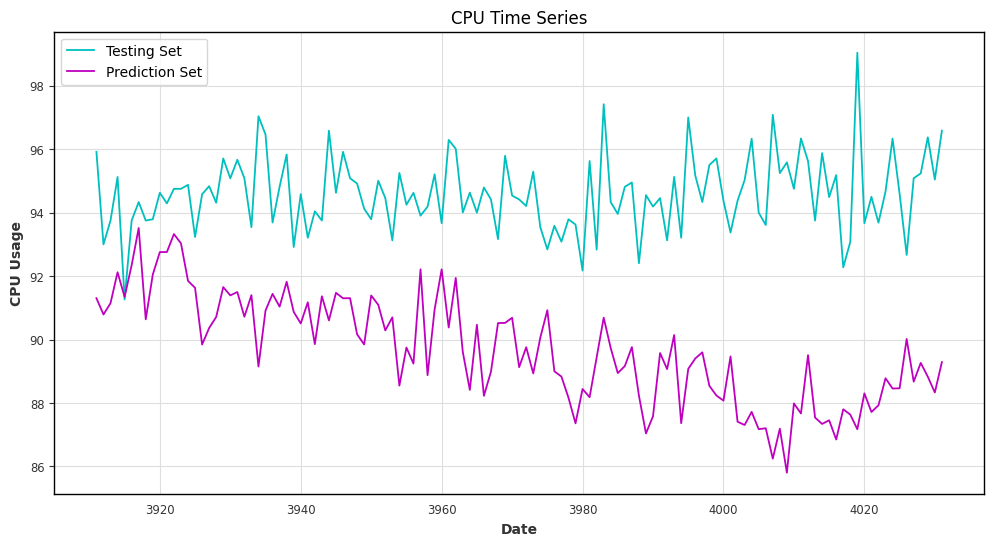}%
\includegraphics[scale=0.3, trim={20pt 5pt 0pt 20pt}, clip]{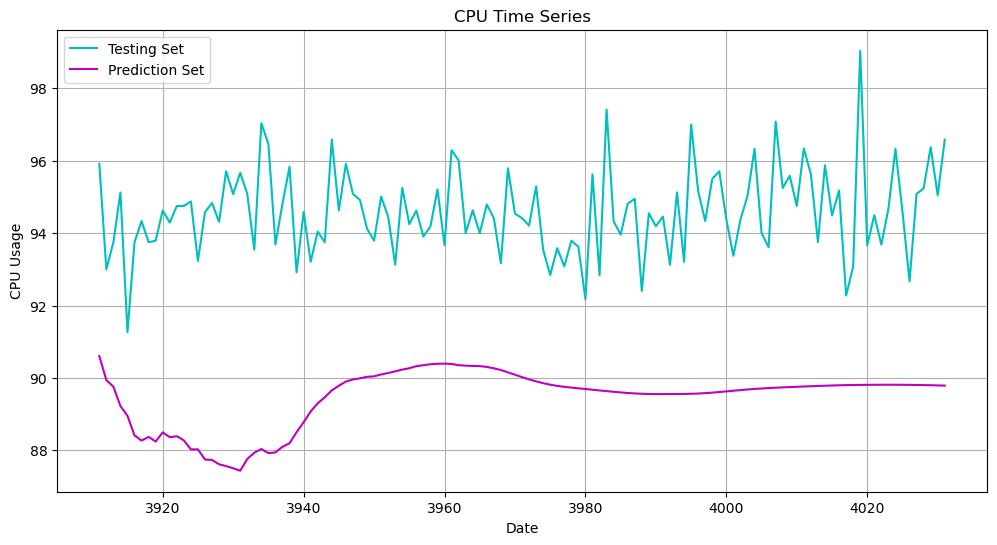}%
\caption{N-HiTS and DFFNN}
\end{subfigure}
\begin{subfigure}[b]{1\linewidth}
\lineskip=0pt
\includegraphics[scale=0.3, trim={20pt 5pt 0pt 20pt}, clip]{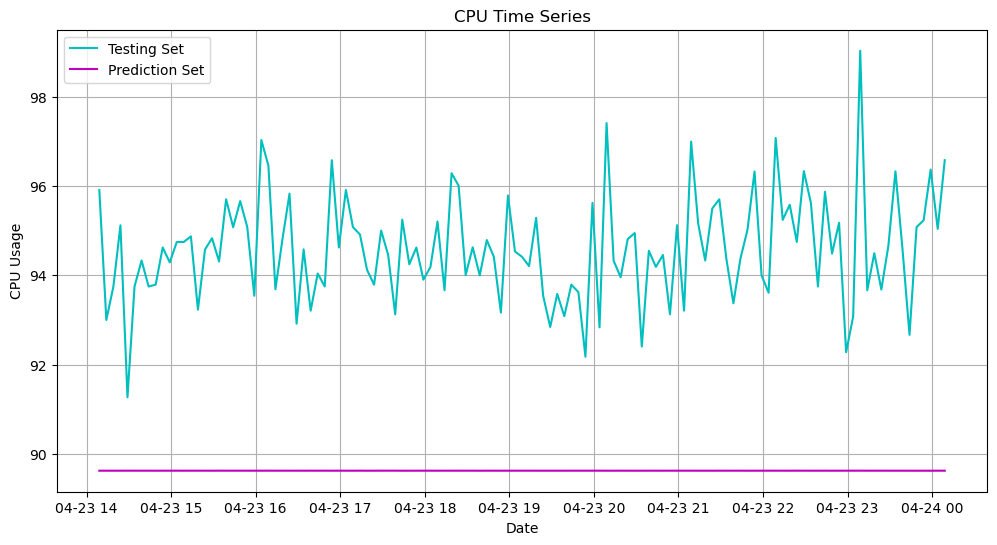}%
\includegraphics[scale=0.3, trim={5pt 5pt 0pt 20pt}, clip]{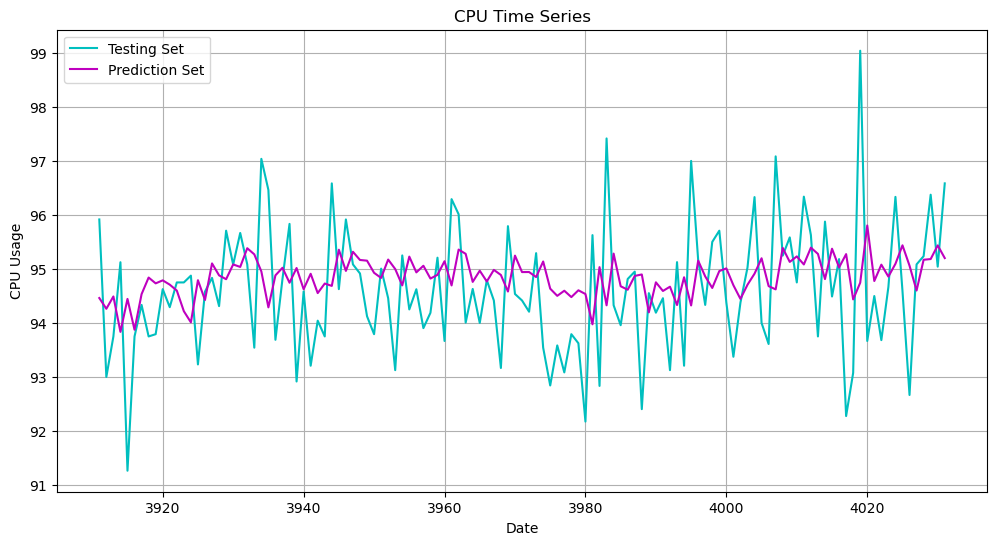}%
\caption{BNN and CNN.}
\end{subfigure}
\rule{\textwidth}{0.1pt}
\caption{Continue.}\label{CPUResults}
\end{figure}
\clearpage
\begin{figure}[t]
\ContinuedFloat
\begin{subfigure}[b]{1\linewidth}
\lineskip=0pt
\includegraphics[scale=0.3, trim={20pt 5pt 0pt 20pt}, clip]{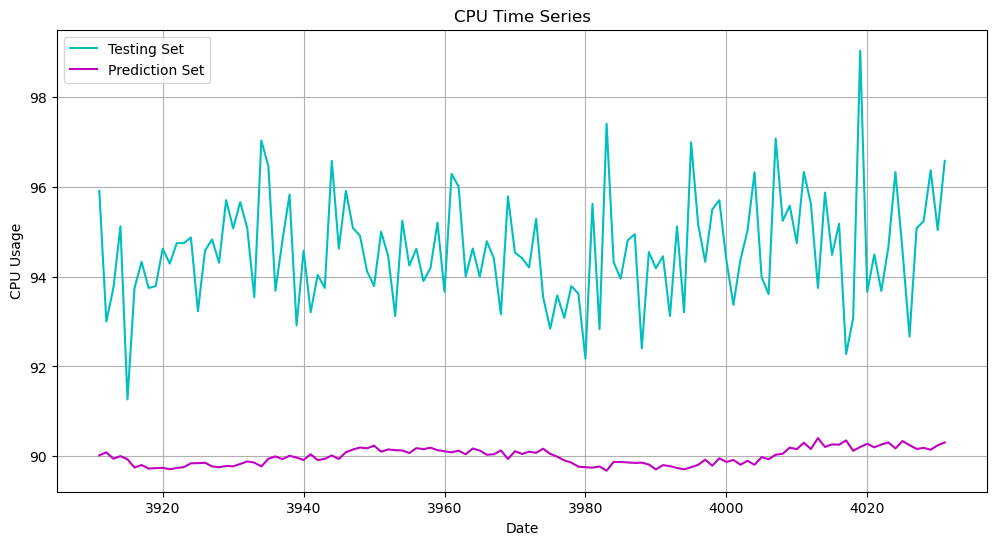}%
\includegraphics[scale=0.3, trim={5pt 5pt 0pt 20pt}, clip]{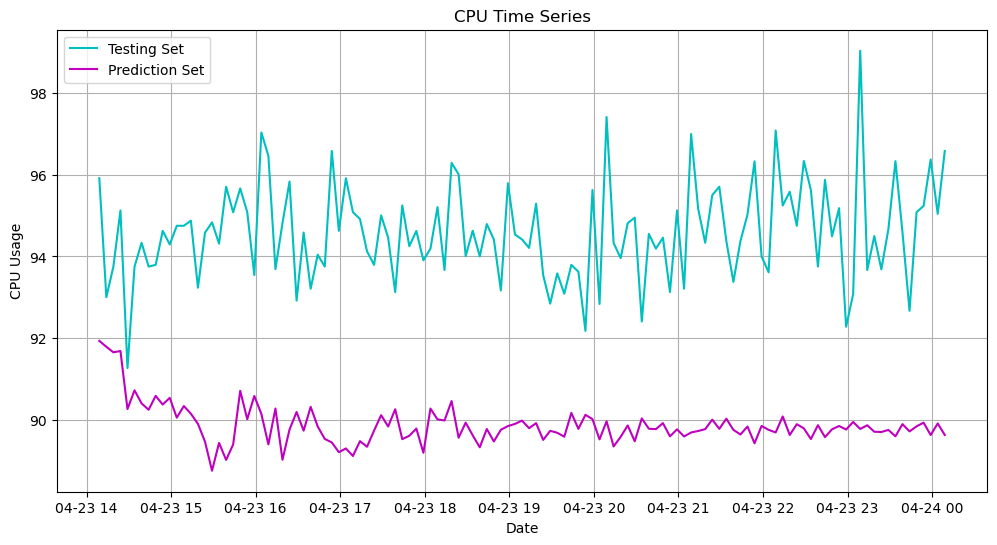}%
\caption{TCN and STFT}
\end{subfigure}
\begin{subfigure}[b]{1\linewidth}
\lineskip=0pt
\includegraphics[scale=0.3, trim={5pt 5pt 0pt 20pt}, clip]{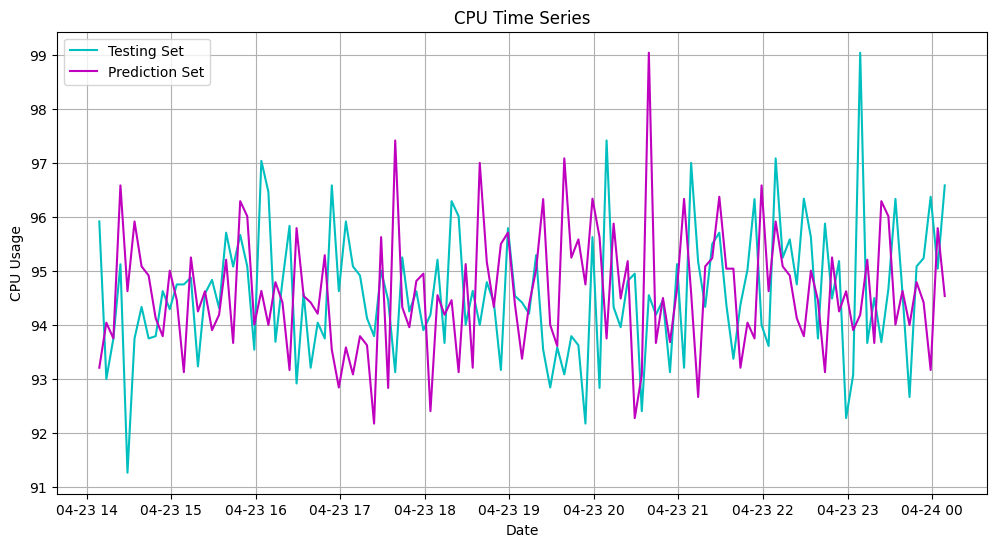}%
\includegraphics[scale=0.26, trim={20pt 5pt 0pt 20pt}, clip]{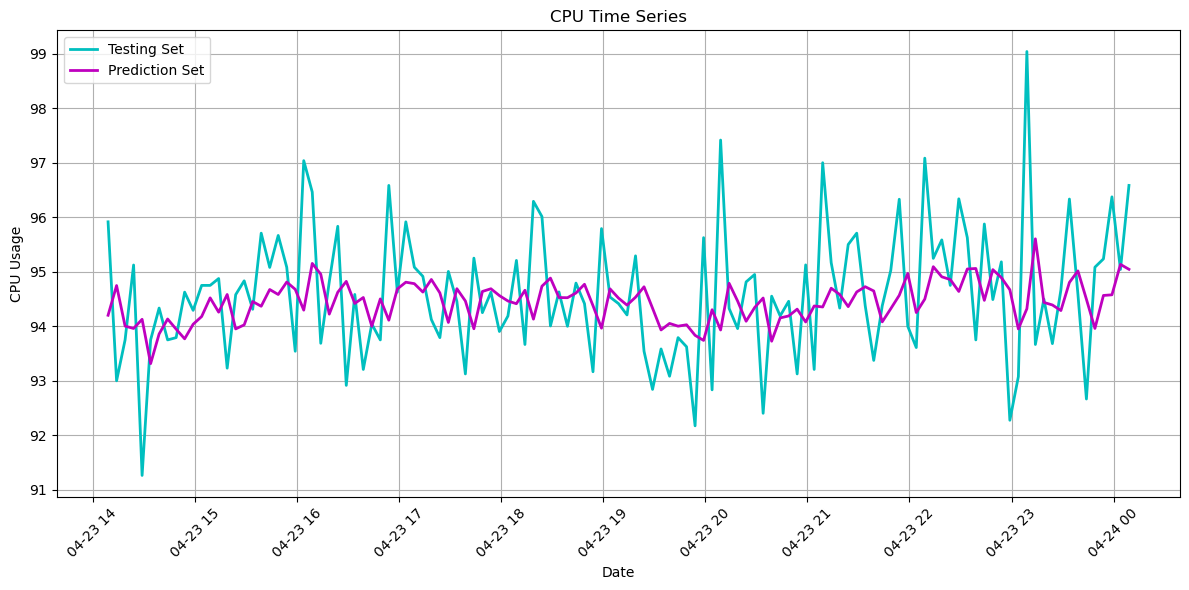}%
\caption{RL and GNN.}
\end{subfigure}
\rule{\textwidth}{0.1pt}
\caption{Prediction results for CPU TS.}\label{CPUResults}
\end{figure}

For the ML purpose, we employ a window size of $30$ to generate the input features from TS data. Each window is used as an input for our models, allowing them to learn patterns and make predictions based on a set number of previous values.
The implementation of some algorithms can be significantly time-consuming, such as RNN, TFT, and BNN. In contrast, some algorithms, like tree-based ensemble methods, Prophet, N-BEATS, N-HiTS, RL, and GNN, are notably faster. The speed of DFFNN, CNN, and TCN depends on various factors, including the number of HLs, the number of neurons and filters, and the size of the dataset. However, the models we consider here are relatively fast.
Some algorithms, like RNNs, can effectively predict initial points, while others, such as DeepAR, tree-based ensemble methods, N-BEATS, CNN, TCN, and GNN, have the capability to make accurate predictions for the more distant future as well.
\subsection{Sunspot TS}
Sunspots from \cite{robervaltsunspots} are cooler areas on the surface of the Sun that are created by massive changes in the Sun's magnetic field. This dataset includes monthly mean of total sunspot number from $1749$ to $2021$, averaging about $273$ years. These variations in the Sun's magnetic field can lead to solar flares and coronal mass ejections, which can have significant effects on Earth's atmosphere and climate.
We divide the data into three sets: training, validation, and testing, as shown in Figure \ref{SunspotPlot}, with the resulting predictions displayed in Figure \ref{SunResults}. Each prediction plot row is labelled with subcaptions to indicate which algorithm it corresponds to.

Although we do not observe any significant trends in the data, our experiments indicate that implementing first-order differencing is beneficial. Additionally, we apply a $240$th-order differencing, followed by normalizing the TS data based on the training statistics.
Also, the selected ARIMA model has a high computational cost and produces inappropriate predictions. It performs ineffectively, particularly for the initial data points. One reason is that the TS is recorded over centuries, making it difficult for the ARIMA model to accurately predict future values.

The best RNN predictions are for DeepAR and LSTM. Considering the time consuming and accuracy, the DeepAR acts acceptable in forecasting. We experiment with different RNNs using the MSE loss function and various hyper-parameters.
Two HLs with $50$ neurons and tanh activation are used in the Elman, followed by one Dense Layer (DeLa).
In the LSTM, three HLs with $40$, $30$, and $20$ neurons and ReLU activation are applied, followed by global max pooling and one DeLa.
For the GRU, two HLs with $50$ and $30$ neurons are utilized, both utilizing tanh activation, followed by one DeLa.
The bidirectional LSTM has two HLs with $50$ and $30$ neurons, using tanh activation, followed by one DeLa.
The deep Elman has five HLs with 30 neurons each using tanh activation, followed by one DeLa.
All algorithms utilize Adam optimizer, except for the LSTM, which employs the Adamax variant. Adam and Adamax have slight differences; Adam utilizes the L2 norm for scaling, whereas Adamax employs the infinity norm, which provides more stable updates when gradients are large or sparse.

Tree-based ensemble algorithms act exceptional predictive performance across all our datasets, particularly in terms of accuracy and evaluation metrics.
The lagged differences ranging from $1$ to $30$ are applied for the RF, XGBoost, and LightGBM models.
For the LightGBM, we add additional features, including the rolling mean with windows of $3$ and $6$, as well as the rolling standard deviation with a window of $3$.
Unlike traditional methods, RF, XGBoost, and LightGBM do not require a specified number of epochs for training. Instead, they utilize the number of trees to make decisions. While the accuracy of predictions improves with an increasing number of trees, this also raises the risk of over-fitting.

\begin{figure}[t] 
\begin{subfigure}[b]{1\linewidth}
\lineskip=0pt
\includegraphics[scale=0.6, trim={5pt 5pt 0pt 20pt}, clip]{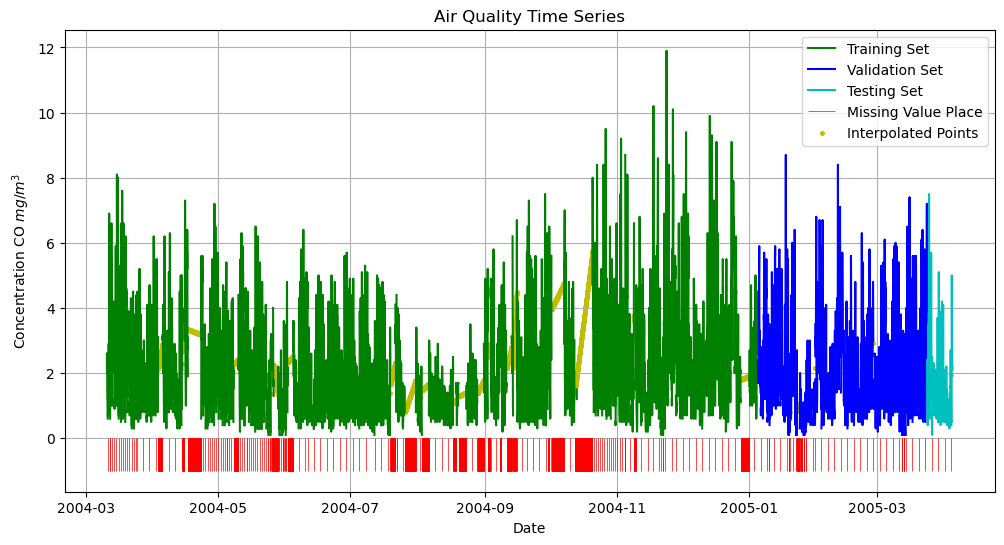}%
\end{subfigure}
\rule{\textwidth}{0.1pt}
\caption{CO Concentration TS. The sizes of the training, validation, and testing sets are $7204$, $1872$, and $281$ samples, respectively. There are $1683$ missing data points in the dataset.}\label{COPlot}
\end{figure}
\begin{figure}[b]
\begin{subfigure}[b]{1\linewidth}
\lineskip=0pt
\includegraphics[scale=0.3, trim={5pt 5pt 0pt 20pt}, clip]{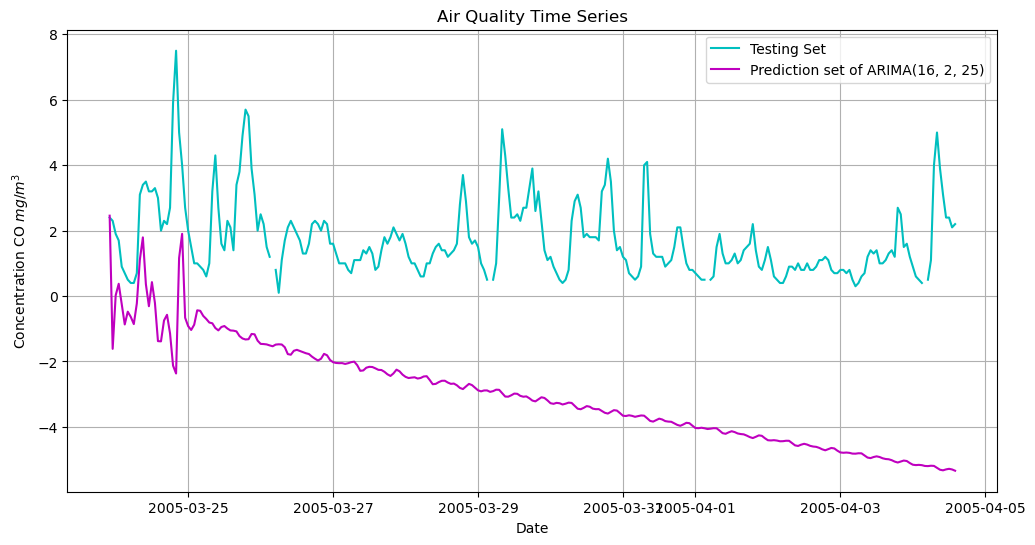}%
\includegraphics[scale=0.3, trim={20pt 5pt 0pt 20pt}, clip]{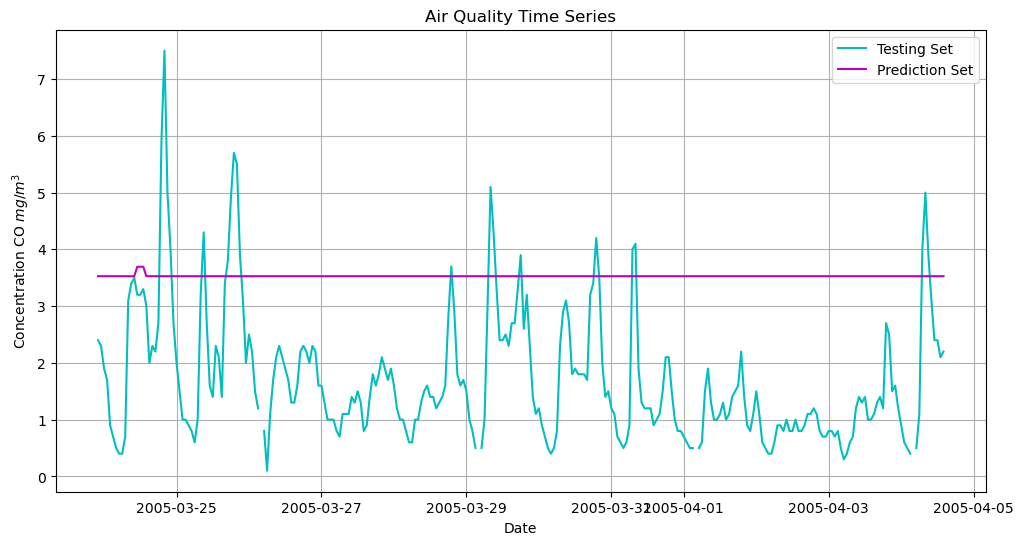}%
\caption{ARIMA and Elman.}
\end{subfigure}
\rule{\textwidth}{0.1pt}
\caption{Continue.}\label{COResults}
\end{figure}
\begin{figure}[b]
\ContinuedFloat
\begin{subfigure}[b]{1\linewidth}
\lineskip=0pt
\includegraphics[scale=0.3, trim={5pt 5pt 0pt 20pt}, clip]{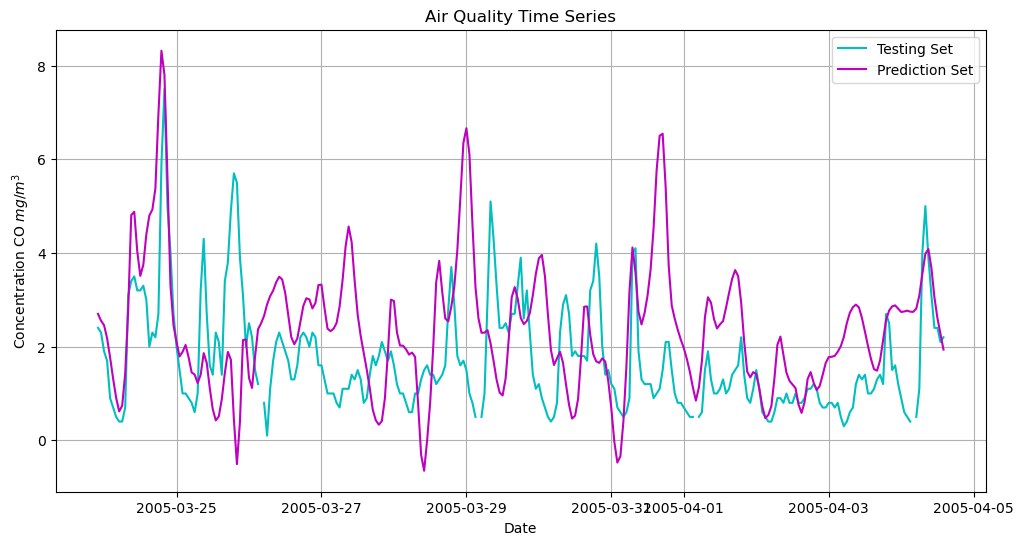}%
\includegraphics[scale=0.3, trim={20pt 5pt 0pt 20pt}, clip]{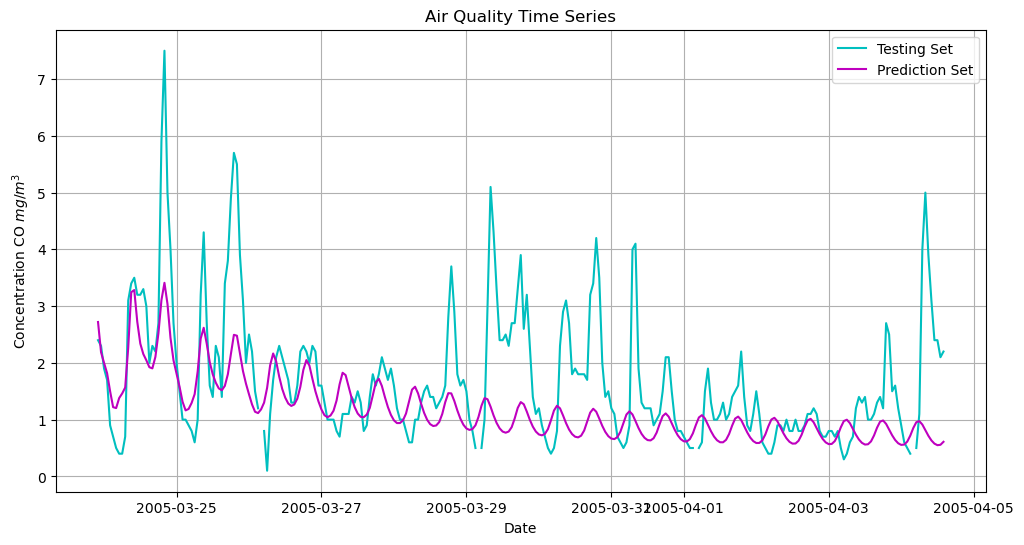}%
\caption{LSTM and GRU.}
\end{subfigure}
\begin{subfigure}[b]{1\linewidth}
\lineskip=0pt
\includegraphics[scale=0.3, trim={5pt 5pt 0pt 20pt}, clip]{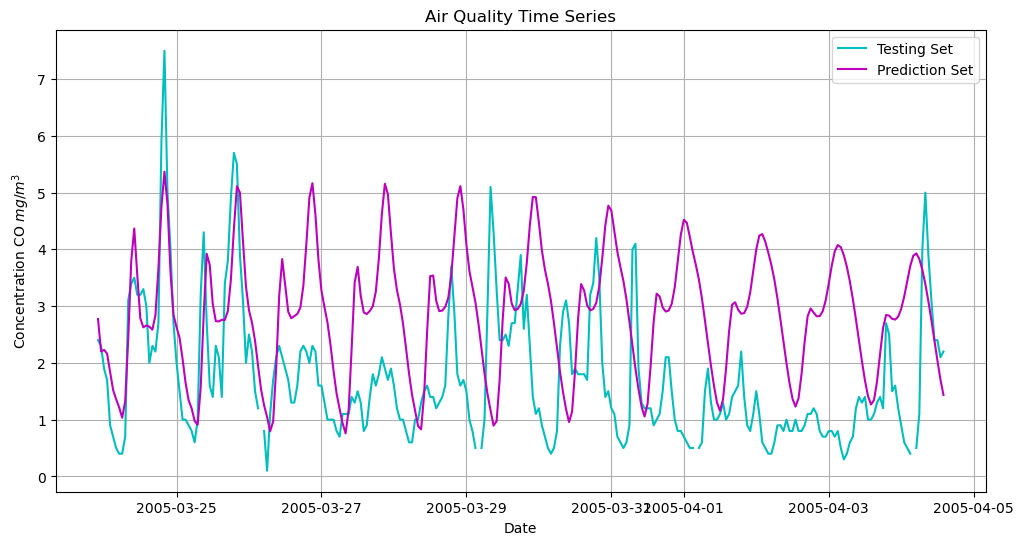}%
\includegraphics[scale=0.3, trim={20pt 5pt 0pt 20pt}, clip]{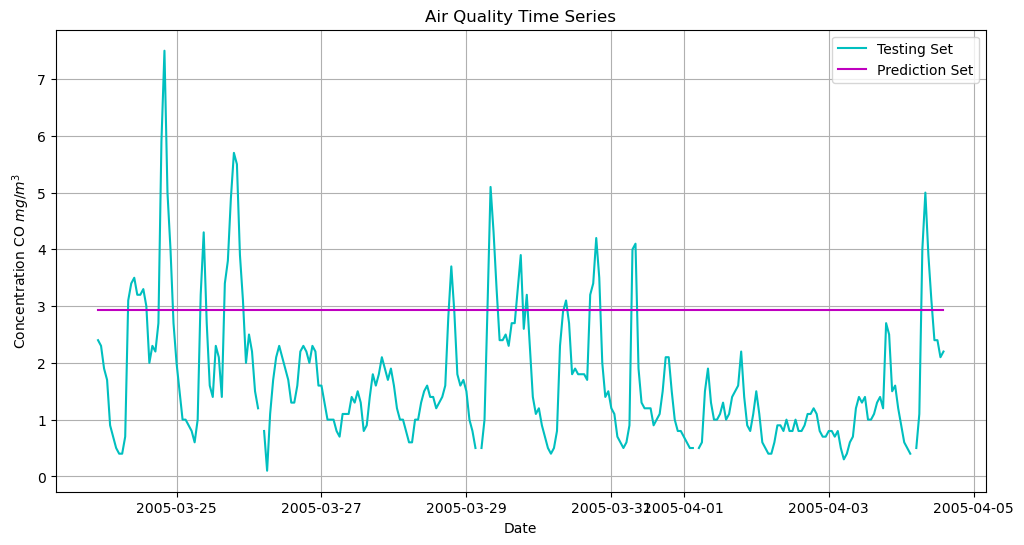}%
\caption{Bidirectional LSTM and Deep Elman.}
\end{subfigure}
\begin{subfigure}[b]{1\linewidth}
\lineskip=0pt
\includegraphics[scale=0.3, trim={5pt 5pt 0pt 20pt}, clip]{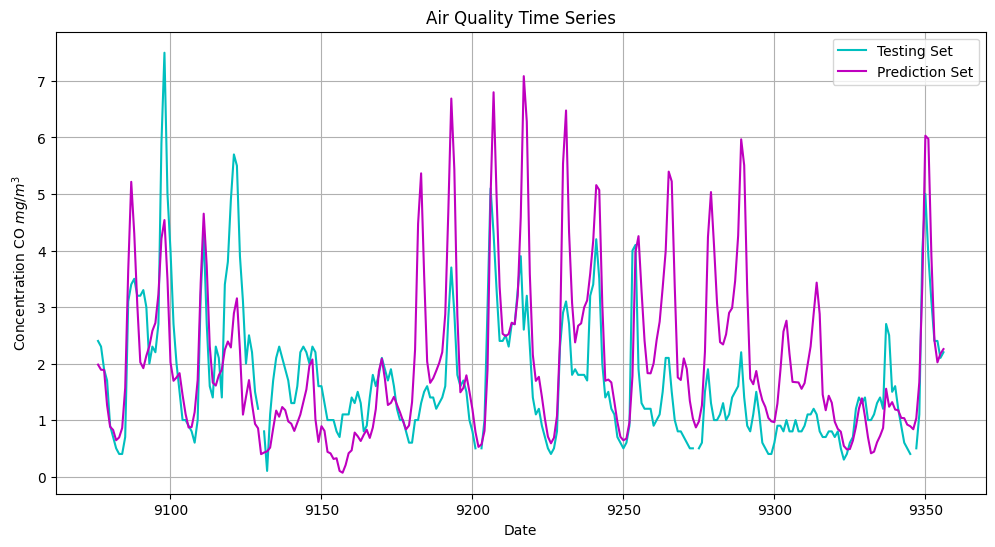}%
\includegraphics[scale=0.3, trim={20pt 5pt 0pt 20pt}, clip]{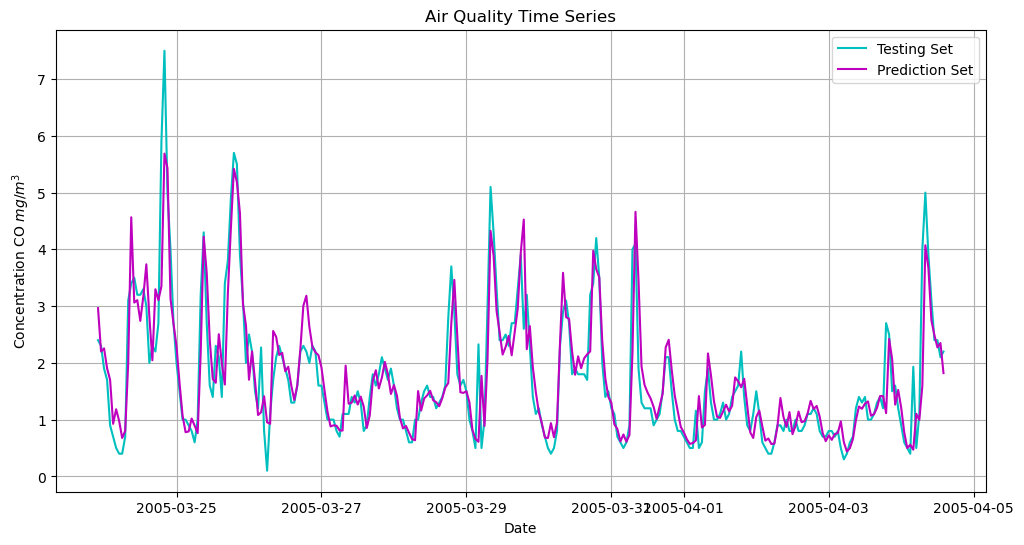}%
\caption{DeepAR and RF.}
\end{subfigure}
\begin{subfigure}[b]{1\linewidth}
\lineskip=0pt
\includegraphics[scale=0.3, trim={20pt 5pt 0pt 20pt}, clip]{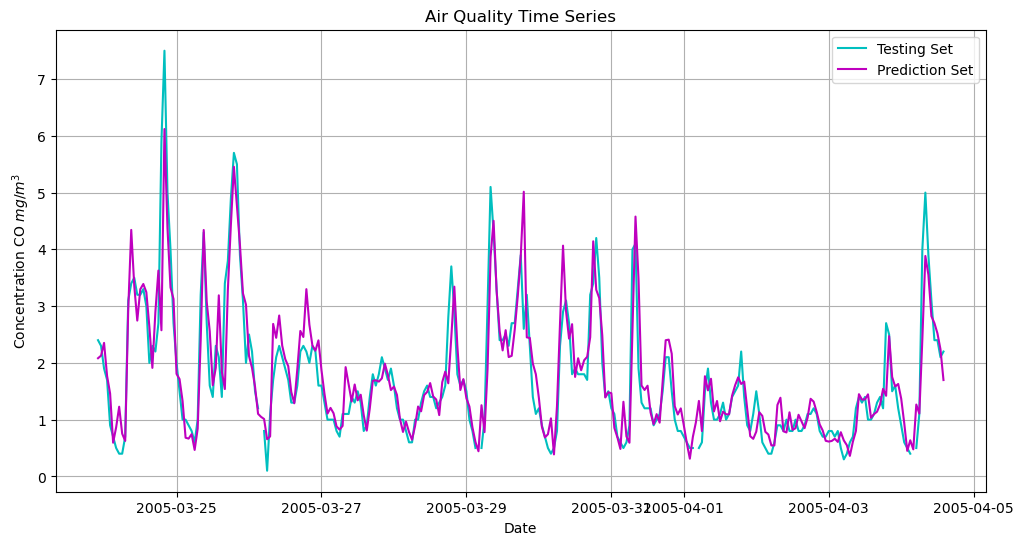}%
\includegraphics[scale=0.3, trim={5pt 5pt 0pt 20pt}, clip]{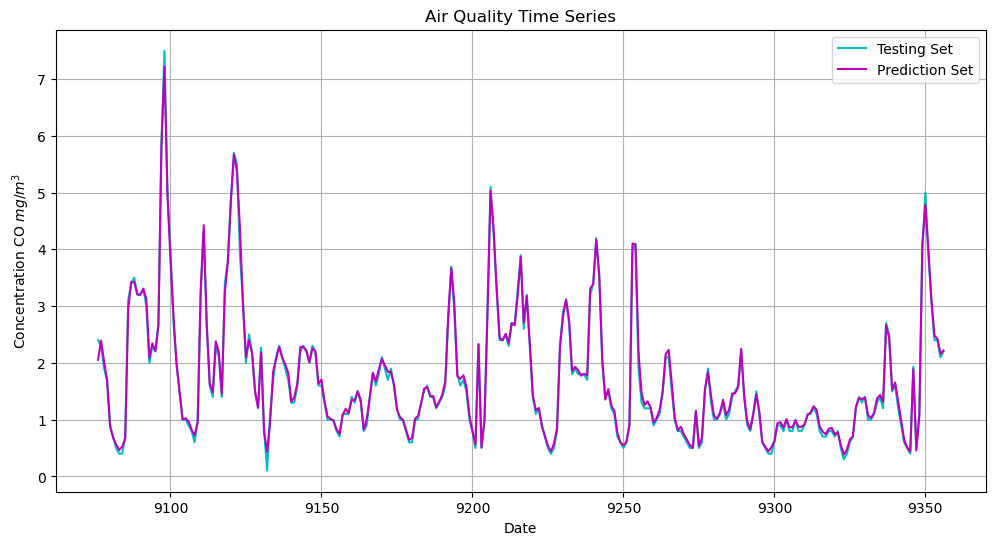}%
\caption{XGBoost and LightGBM.}
\end{subfigure}
\rule{\textwidth}{0.1pt}
\caption{Continue.}\label{COResults}
\end{figure}
\begin{figure}[t]
\ContinuedFloat
\begin{subfigure}[b]{1\linewidth}
\lineskip=0pt
\includegraphics[scale=0.3, trim={20pt 5pt 0pt 20pt}, clip]{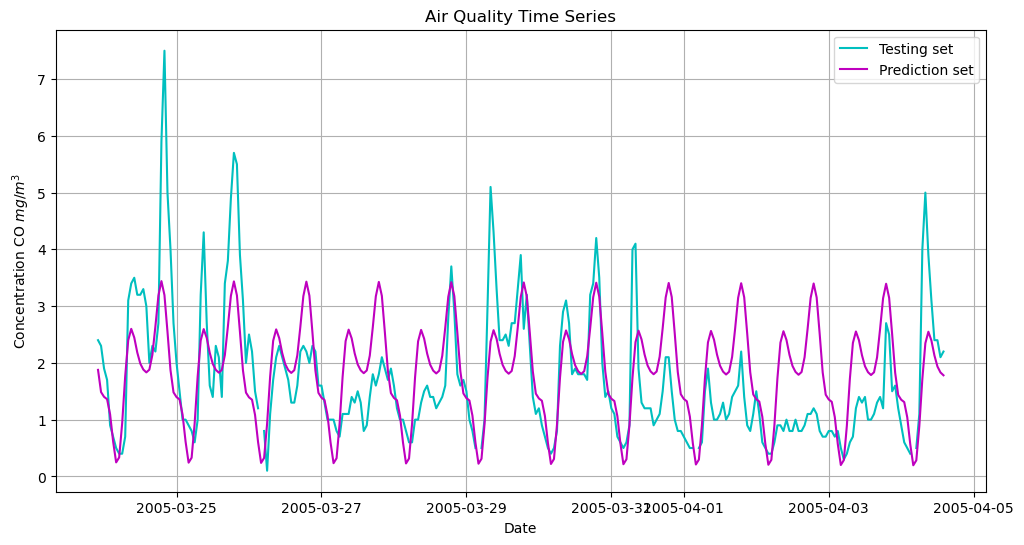}%
\includegraphics[scale=0.3, trim={20pt 5pt 0pt 20pt}, clip]{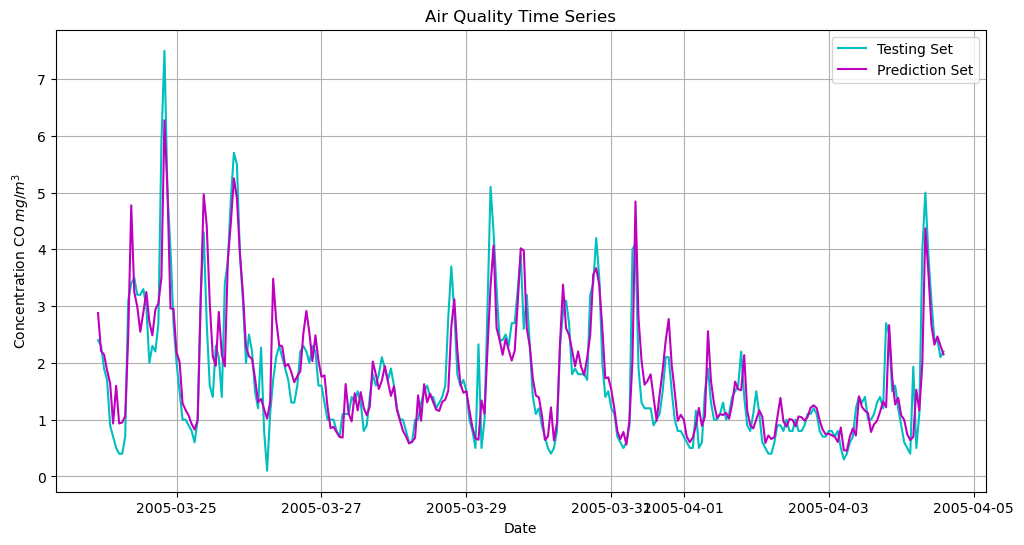}%
\caption{Prophet and WBT.}
\end{subfigure}
\begin{subfigure}[b]{1\linewidth}
\lineskip=0pt
\includegraphics[scale=0.3, trim={5pt 5pt 0pt 20pt}, clip]{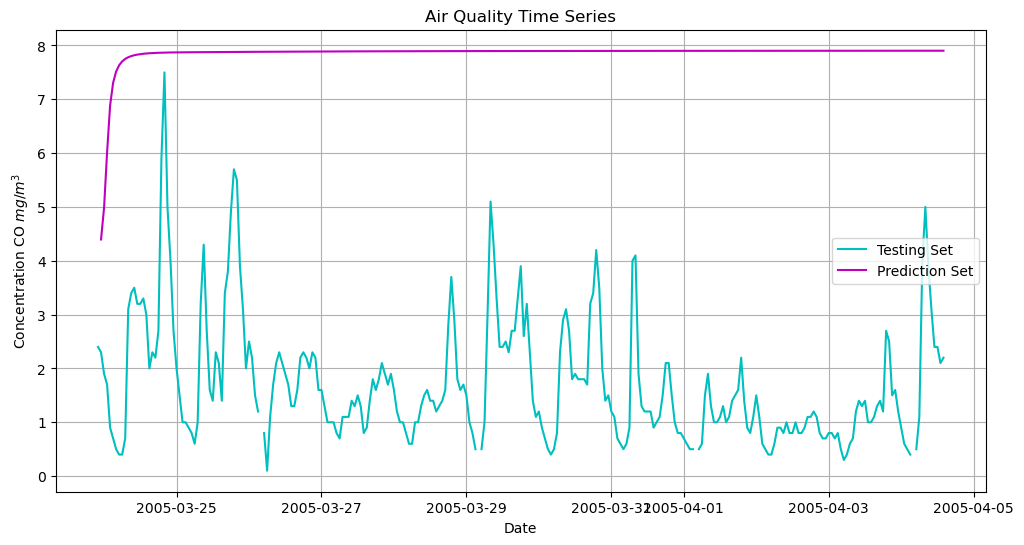}%
\includegraphics[scale=0.3, trim={20pt 5pt 0pt 20pt}, clip]{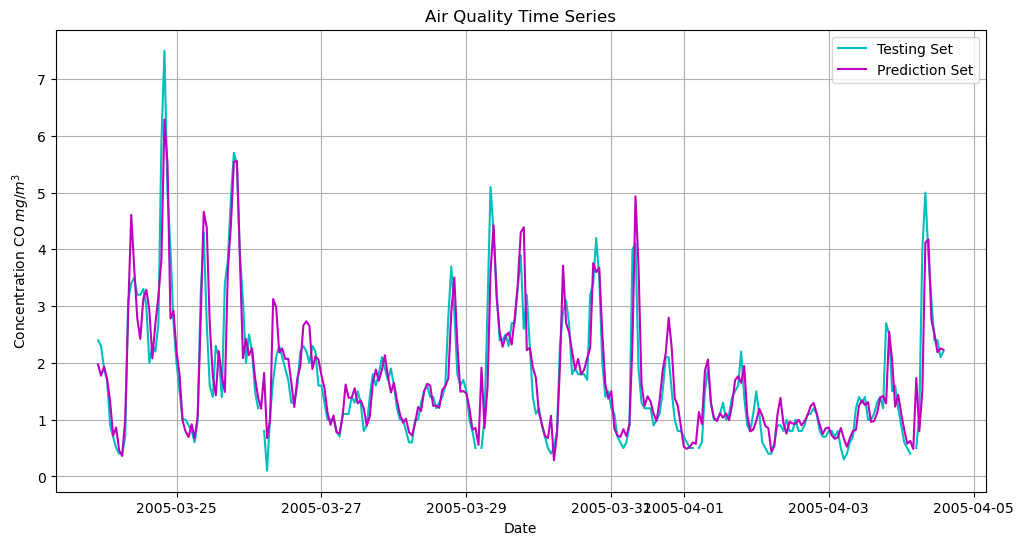}%
\caption{TFT and N-BEATS.}
\end{subfigure}
\begin{subfigure}[b]{1\linewidth}
\lineskip=0pt
\includegraphics[scale=0.3, trim={5pt 5pt 0pt 20pt}, clip]{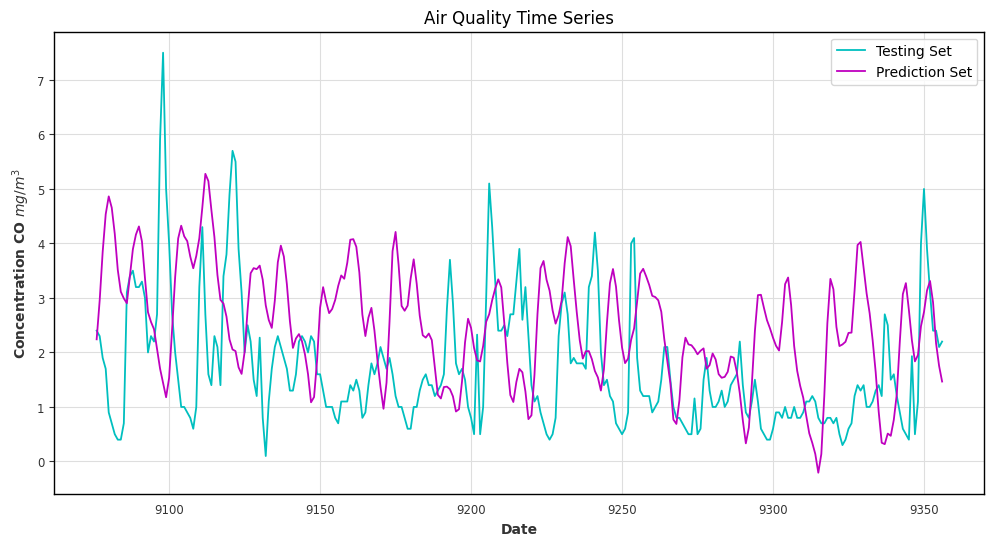}%
\includegraphics[scale=0.3, trim={20pt 5pt 0pt 20pt}, clip]{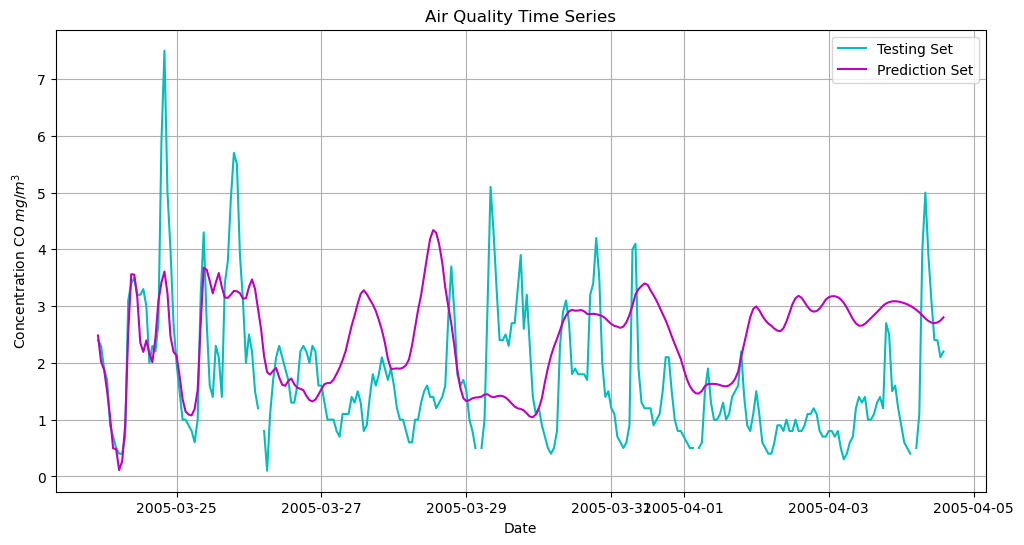}%
\caption{N-HiTS and DFFNN.}
\end{subfigure}
\begin{subfigure}[b]{1\linewidth}
\lineskip=0pt
\includegraphics[scale=0.3, trim={20pt 5pt 0pt 20pt}, clip]{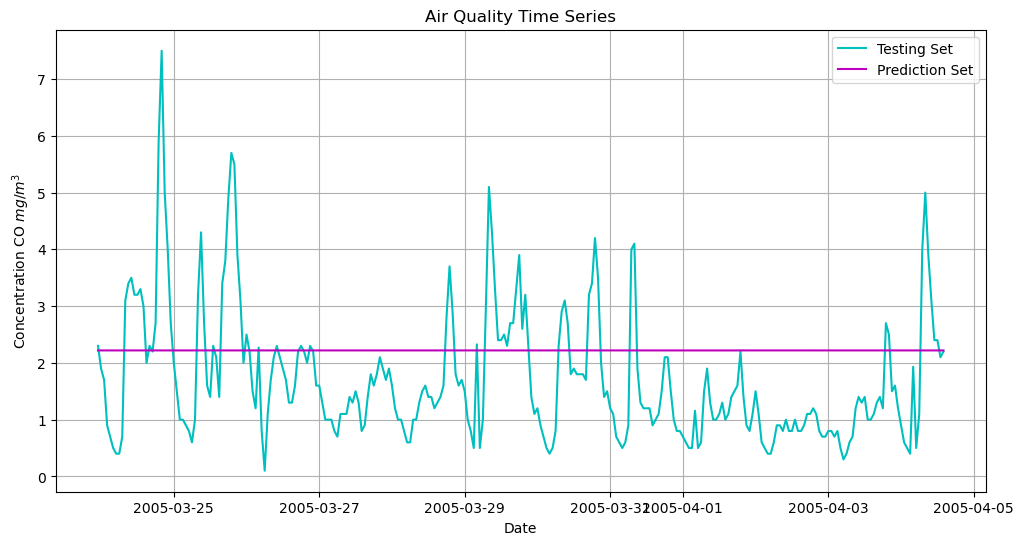}%
\includegraphics[scale=0.3, trim={5pt 5pt 0pt 20pt}, clip]{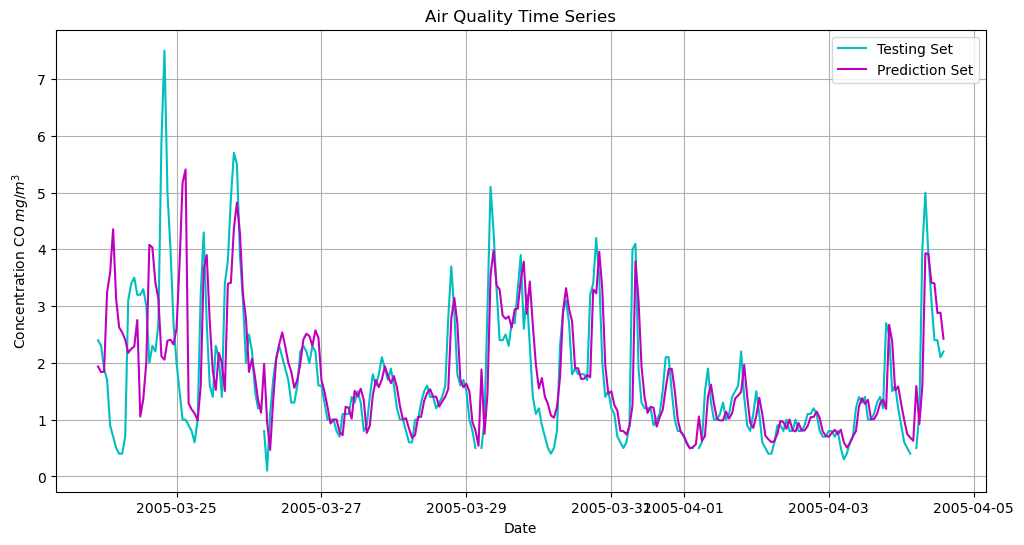}%
\caption{BNN and CNN.}
\end{subfigure}
\rule{\textwidth}{0.1pt}
\caption{Continue.}\label{COResults}
\end{figure}
\clearpage
\begin{figure}[t]
\ContinuedFloat
\begin{subfigure}[b]{1\linewidth}
\lineskip=0pt
\includegraphics[scale=0.3, trim={20pt 5pt 0pt 20pt}, clip]{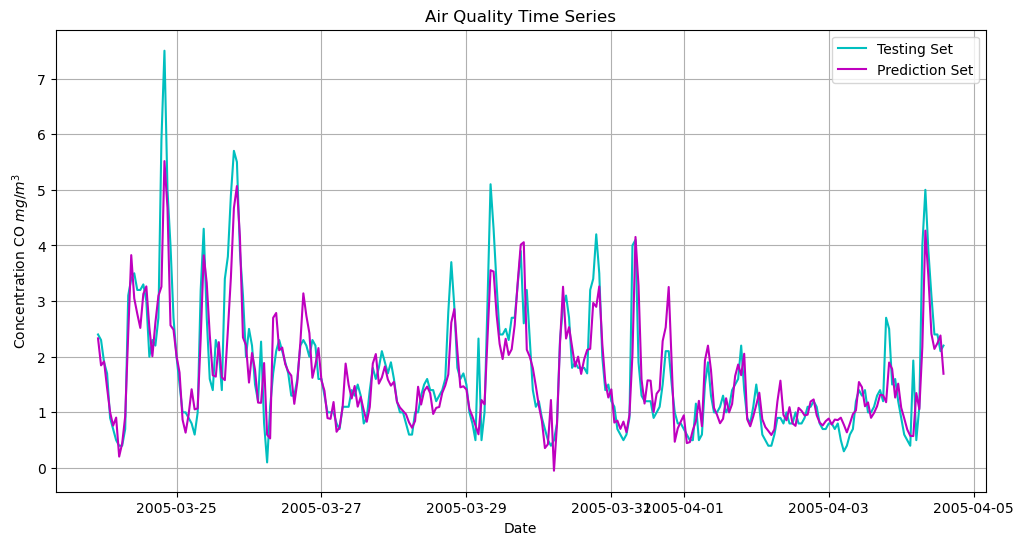}%
\includegraphics[scale=0.3, trim={5pt 5pt 0pt 20pt}, clip]{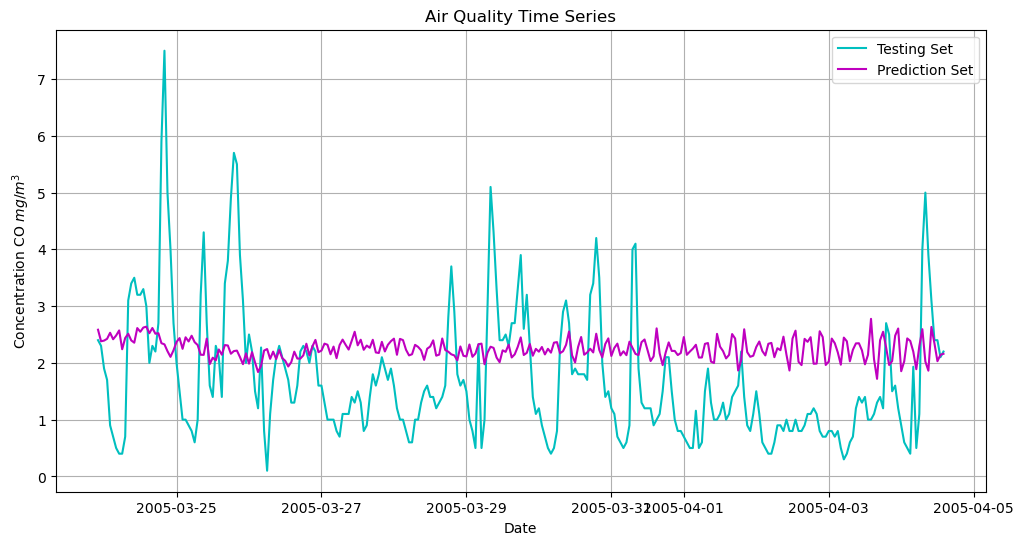}%
\caption{TCN and STFT}
\end{subfigure}
\begin{subfigure}[b]{1\linewidth}
\lineskip=0pt
\includegraphics[scale=0.3, trim={5pt 5pt 0pt 20pt}, clip]{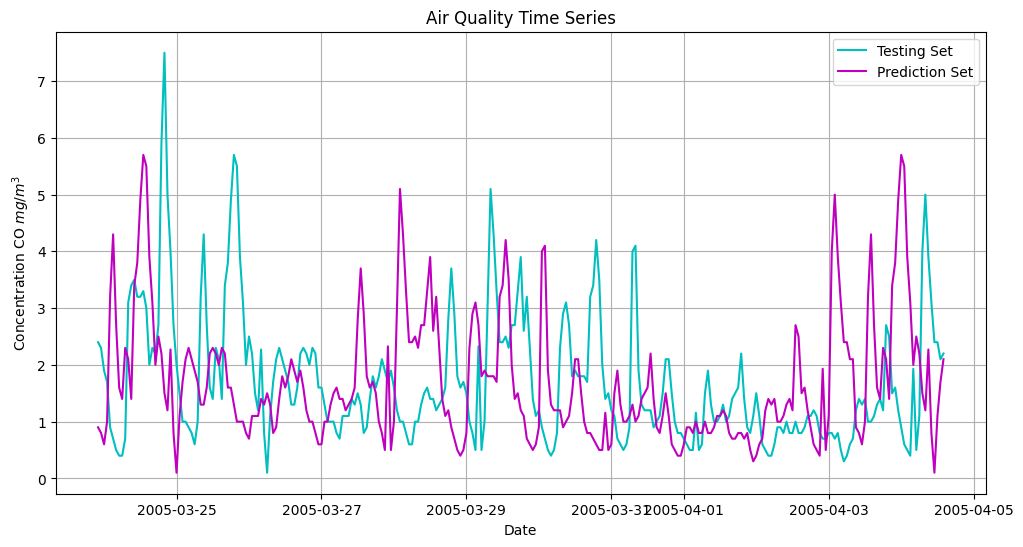}%
\includegraphics[scale=0.26, trim={20pt 5pt 0pt 20pt}, clip]{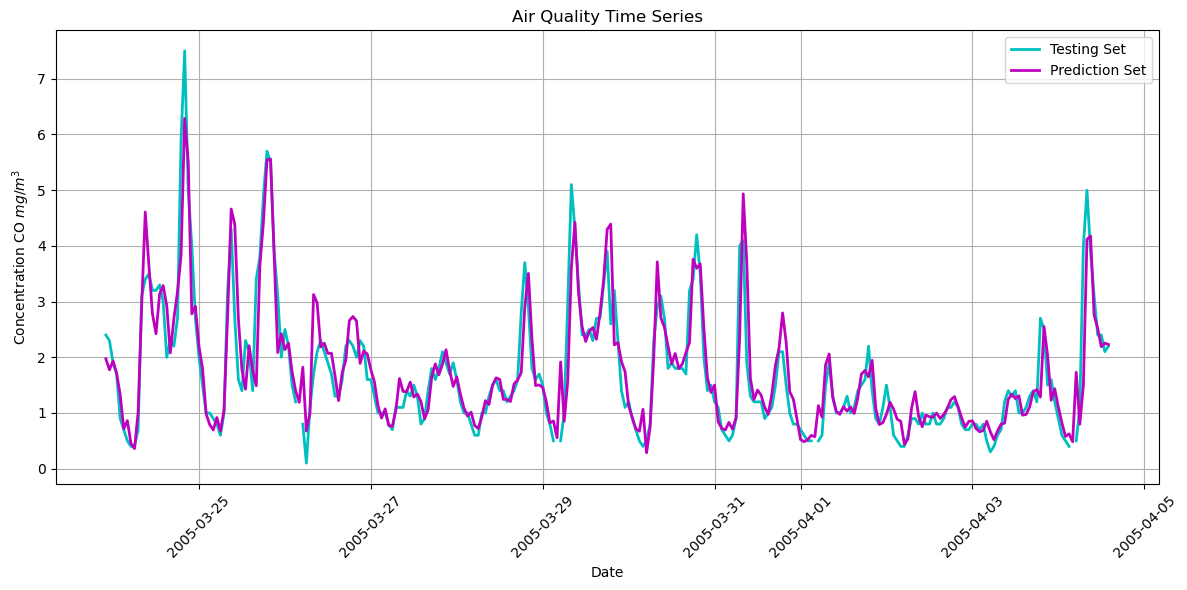}%
\caption{RL and GNN.}
\end{subfigure}
\rule{\textwidth}{0.1pt}
\caption{Prediction results for air quality TS.}\label{COResults}
\end{figure}

The Haar functions at level $5$ are used to extract the WBT features. These features are then incorporated into a RF model, along with lagged differences ranging from $1$ to $30$.
We employ a multiplicative Prophet model with $30$ MCMC samples across $4$ chains.
For the TFT, we apply differencing of lags $1$ to $30$, along with a rolling mean and standard deviation using a window size of $30$.
The best model among the specialized TS models is WBT and N-BEATS.
The prediction curves roughly resemble the structure of the testing set.

In the DFFNN, two HLs containing $50$ neurons each, activated by the ReLU function, are followed by one DeLa utilizing the tanh activation function.
Weights, biases, and observations in the BNN are modeled using a normal distribution with ten HLs. To enhance the complexity of the BNN model, we incorporate lagged differencing features with ranges from $1$ to $30$.
The CNN has two one-dimensional convolutional layers with $32$ and $64$ filters followed by the ReLU activation. This is followed by global average pooling and a DeLa.
The TCN has two main convolutional HLs, each with $64$ filters, a kernel size of $2$, and ReLU activation. Each layer is followed by a dropout rate of $0.2$. Subsequently, there are a flattening layer and a DeLa.

The STFT features are utilized within a GRU architecture comprising three HLs, containing $40$, $30$, and $20$ neurons, respectively. This architecture is followed by a dropout rate of $0.1$ and a DeLa.

The architecture of the GNN consists of two graph convolutional layers, with the first layer expanding the initial single node feature into $16$ features followed by a ReLU activation, and the second layer reducing the dimensionality back to a single output feature per node.
The remaining algorithms have been implemented as usual, and we do not define any special characteristics.
As a consequence, the most accurate predictions, ranked by the lowest error rates, are as follows: LightGBM, RF, XGBoost, N-BEATS, WBT, TCN, GNN, CNN, and DeepAR.

\subsection{CPU Usage TS with Outliers}
We select the real CPU utilization TS data from \cite{numentanab}, network bytes collected by Amazon Watch Server Cloud metrics. The data was recorded every $5$ minutes from April $10$ to $24$, $2014$.
This dataset contains outliers, as shown in Figure \ref{CPUPlot}.
Our goal is to predict the test set using models trained on the training and validation sets, which contain outliers. We want to determine whether the presence of this data affects the prediction results. Some algorithms perform well, so the preparation of outliers is not necessary. Consequently, they facilitate an easy prediction process with high accuracy.

For this example, first and twelfth-order differencing are performed, followed by normalization for the ARIMA model. This TS exhibits persistent patterns, making it challenging to predict using ARIMA.
Also, we apply similar ML procedures to the sunspot data, with a few exceptions. The LSTM architecture consists of two HLs containing $20$ and $10$ neurons, followed by a global max pooling layer and a DeLa. The DFFNN has three HLs with $50$, $40$, and $30$ neurons, along with a DeLa.
The forecasting plots are presented in Figure \ref{CPUResults}.
The LightGBM model demonstrates excellent predictive accuracy, even in the presence of outliers, while the WBT, N-BEATS, RF, XGBoost, CNN, RL, GNN, and DeepAR models provide acceptable predictions, ranked respectively in terms of accuracy and error minimization.
Algorithms such as bidirectional LSTM, LSTM, GRU, and Prophet exhibit low error rates; however, their predictions may not be suitable and have no information.
\subsection{Air CO Concentration TS with Missing Data}
For this subsection, we analyze the real hourly averaged concentration of carbon monoxide (CO) in $mg/m^3$, as displayed in Figure \ref{COPlot}, adapted from \cite{uciairquality}.
The sensor was located in an Italian city's significantly polluted area at road level.
Data were recorded from March $2004$ to February $2005$ from air quality chemical sensor devices deployed in the field.
This dataset contains missing information.
The employed filling method is first-order spline interpolation, as higher-order splines produced negative values, which are not acceptable for a positive variable.

The same preparations to the CPU TS are done for predicting with ARIMA model.
Furthermore, we apply similar ML algorithms to the sunspot TS, with the exception of the Prophet algorithm, as shown the predictions in Figure \ref{COResults}.
The additive Prophet model performs effectively on this data without the need for MCMC sampling.
The algorithms that accurately predicted the entire testing set, ranked by the lowest error rates, are LightGBM, RF, WBT, XGBoost, TCN, GNN, CNN, Prophet, GRU, DeepAR, DFFNN, RL, LSTM, bidirectional LSTM, N-BEATS.
Some algorithms like Prophet, GRU, DFFNN, and bidirectional LSTM provide suitable predictions for the initial data points; however, they struggle to predict long-term future values.
\section{Conclusion}\label{S8}
In this paper, we conduct a comprehensive review of ML algorithms suitable for TS forecasting. We provide a brief explanation of each algorithm. Ultimately, we selected three TS datasets for our analysis: the first dataset consisted of complete data collected over centuries, the second contained outliers, and the third included gaps in the data. We implement the introduced algorithms on these datasets to perform predictions on unobserved testing set.

We determine the hyper-parameters for the algorithms based on our experience and repeated implementations of the algorithms.
In our findings, we note that the LightGBM algorithm outperforms the others, yielding highly accurate predictions for all three datasets. This algorithm also demonstrates a high execution speed, which facilitates data analysis.

This paper is suited for individuals working with TS data, providing guidance on selecting an appropriate method for their data analysis needs.
In our analysis, tree-based ensembles, N-BEATS, CNN, TCN, GNN, DeepAR, and RL algorithms demonstrate strong predictive performance on testing sets including outliers and missing data. This outcome does not imply that other algorithms are unsuitable; rather, it indicates that they require further attention to hyper-parameters and feature engineering to perform effectively.

\section*{\bf Declarations}
This work was supported by the Innovative Research Group Project of the National Natural Science Foundation of China (Grant No. 51975253). The authors report no conflicts of interest or personal ties that may affect the work in this paper. The contributions are as follows:\\
Seyedeh Azadeh Fallah Mortezanejad: Methodology, Conceptualization, Validation, Writing-original draft, Writing-review and editing.\\
Ruochen Wang: Supervision, Methodology, Investigation, Writing-original draft, Writing-review and editing.

\bibliographystyle{amsplain}

\end{document}